\documentclass[dvipsnames,format=sigconf,anonymous=false,authorversion,nonacm,table]{acmart}

\usepackage{color,soul}
\usepackage{lipsum}
\hypersetup{
 colorlinks=true,
}
\usepackage{enumitem}
\usepackage{caption}
\usepackage{subcaption}
\usepackage{cleveref}
\usepackage{xspace}
\usepackage{adjustbox}
\usepackage[detect-weight=true]{siunitx}
\usepackage{multicol,multirow,booktabs,makecell,array}
\usepackage{algorithm}
\usepackage[commentColor=teal,italicComments=false]{algpseudocodex}
\crefname{algocf}{algorithm}{algorithms}
\Crefname{algocf}{Algorithm}{Algorithms}
\usepackage{gensymb}
\usepackage{xcolor}
\usepackage{amsmath}
\usepackage{circledsteps}
\usepackage{stackengine}
\setstackEOL{\\}

\algrenewcommand\algorithmicensure{\textbf{Input:}} 

\definecolor{blue}{rgb}{0.518, 0.741, 0.812}
\definecolor{red}{rgb}{0.871, 0.537, 0.573}
\definecolor{pastelyellow}{rgb}{0.518, 0.678, 0.473}
\definecolor{green}{rgb}{0.937, 0.914, 0.667}

\DeclareMathOperator*{\argmax}{argmax}

\AtBeginDocument{%
 \providecommand\BibTeX{{%
 \normalfont \objB\kern-0.5em{\scshape i\kern-0.25em b}\kern-0.8em\TeX}}}

\colorlet{rejectColor}{pastelyellow!50}

\definecolor{s1color_}{rgb}{0.8745, 0.1255, 0.1294}
\colorlet{s1color}{s1color_!70} 
\definecolor{s2color_}{rgb}{0.4627, 0.6039, 0.8627}
\colorlet{s2color}{s2color_!80}

\begin{document}

\newcommand{\objI}{$\mathcal{I}^{\uparrow}$\xspace}
\newcommand{\objS}{$\mathcal{S}^{\downarrow}$\xspace}
\newcommand{\objC}{$\mathcal{C}^{\uparrow}$\xspace}
\newcommand{\objF}{$\mathcal{F}^{\uparrow}$\xspace}
\newcommand{\objB}{$\mathcal{B}^{\downarrow}$\xspace}
\newcommand{\objT}{$\mathcal{T}^{\downarrow}$\xspace}

\newcommand{\setone}{\Circled[outer color=s1color,fill color=s1color]{1}\xspace}
\newcommand{\settwo}{\Circled[outer color=s2color,fill color=s2color]{2}\xspace}

\newcommand{\STAB}[1]{\begin{tabular}{@{}c@{}}#1\end{tabular}}

\title[Many-Objective Evolutionary Influence Maximization]{Many-Objective Evolutionary Influence Maximization:\\Balancing Spread, Budget, Fairness, and Time}

\newcommand{\algname}{MOEIM\xspace}
\newcommand{\algdesc}{Many-Objective Evolutionary Algorithm for Influence Maximization\xspace}

\author{Elia Cunegatti}
\email{elia.cunegatti@unitn.it}
\affiliation{%
  \institution{University of Trento}
  \city{Trento}
  \country{Italy}
}

\author{Leonardo Lucio Custode}
\email{leonardo.custode@unitn.it}
\affiliation{%
  \institution{University of Trento}
  \city{Trento}
  \country{Italy}
}

\author{Giovanni Iacca}
\email{giovanni.iacca@unitn.it}
\affiliation{%
  \institution{University of Trento}
  \city{Trento}
  \country{Italy}
}

\newcommand\blfootnote[1]{%
  \begingroup
  \renewcommand\thefootnote{}\footnote{#1}%
  \addtocounter{footnote}{-1}%
  \endgroup
}

\renewcommand{\shortauthors}{Cunegatti et al.}

\begin{abstract}
The Influence Maximization (IM) problem seeks to discover the set of nodes in a graph that can spread the information propagation at most. This problem is known to be NP-hard, and it is usually studied by maximizing the influence (\emph{spread}) and, optionally, optimizing a second objective, such as minimizing the \emph{seed set size} or maximizing the influence \emph{fairness}. However, in many practical scenarios multiple aspects of the IM problem must be optimized at the same time. In this work, we propose a first case study where several IM-specific objective functions, namely \emph{budget}, \emph{fairness}, \emph{communities}, and \emph{time}, are optimized on top of the maximization of \emph{influence} and minimization of the \emph{seed set size}. To this aim, we introduce \algname (\algdesc), a Multi-Objective Evolutionary Algorithm (MOEA) based on NSGA-II incorporating graph-aware operators and a smart initialization. We compare \algname in two experimental settings, including a total of nine graph datasets, two heuristic methods, a related MOEA, and a state-of-the-art Deep Learning approach. The experiments show that \algname overall outperforms the competitors in most of the tested many-objective settings. To conclude, we also investigate the correlation between the objectives, leading to novel insights into the topic. The codebase is available at \href{https://github.com/eliacunegatti/MOEIM}{https://github.com/eliacunegatti/MOEIM}. \blfootnote{Corresponding author: {\tt elia.cunegatti@unitn.it}.}
\end{abstract}

\maketitle


\section{Introduction}
\label{sec:intro}

Many real-world problems can be modelled through the interaction between multiple entities. Be it aimed at measuring the effect of political campaigns or commercial advertising, characterizing the connections over a social network such as Facebook or X, the mutual dependencies between parts of a distributed system, the co-purchasing of goods, or the impact of a scientific article over the research community, many such problems can be modelled through a \emph{social network}, namely a graph $\mathcal{G}=(\mathcal{V},\mathcal{E})$ where $\mathcal{V}$ denotes the set of entities in the network (i.e., the vertices, or nodes, of the graph) and $\mathcal{E}$ denotes their connections (the edges). Such graph can be either \emph{directed} or \emph{undirected} (depending on the direction of the information over the edges), \emph{weighted} or \emph{unweighted} (depending on the possibility to quantify a given numerical property of each edge), and \emph{static} or \emph{dynamic} (if nodes and edges are time-variant). 

The specific meaning of entities and connections, obviously, depends on the semantics of the application domain. Nevertheless, one common need in many domains is to identify the most \emph{influential} nodes over the network, i.e., the subset (usually called \emph{seed set}) of nodes from which the \emph{influence} (a concept that is, in turn, application-dependent) can propagate over the highest possible number of other nodes. The search for such seed set is usually referred to as the Influence Maximization (IM) problem, and has been originally defined by Kempe et al. in \cite{kempe2003maximizing}. This problem has been proven to be NP-hard, attracting over the years a massive research effort from the machine learning and optimization community.

The earliest algorithmic attempts at effectively solving the IM problem resorted to greedy heuristics, such as \cite{leskovec2007cost,goyal2011celf++}, with more recent methods combining greedy approaches with statistics-based models \cite{tang2015influence,tang2018online}. Other works, such as \cite{bucur2016influence,lotf2022improved,chawla2023neighbor}, focused on the use of Evolutionary Algorithms (EAs). More recently, Deep Learning (DL) methods, e.g., based on deep Reinforcement Learning (RL) \cite{li2022piano,chen2023touplegdd}, Learnable Embeddings \cite{panagopoulos2020multi,ling2023deep}, or combinations of deep RL and EAs \cite{ma2022influence}, have also been proposed. 

An important limitation in the current state of the art is the fact that, apart from a few works that we will discuss in detail later, most research in the field focuses on the standard, single-objective IM problem formulation from \cite{kempe2003maximizing}, where the only goal is to maximize the number of influenced nodes. On the other hand, various other aspects of the IM problem can be of interest in practical applications, such as minimizing the \emph{seed set size}, as well as achieving \emph{fairness} (either at the beginning, or the end of the influence propagation) across the communities in the network, or optimizing the \emph{time} and \emph{budget} (also called \emph{seed cost}) needed to propagate the influence.

The only works that tried to handle multiple objectives in the IM formulation, e.g., influence and seed set size \cite{bucur2017multi,bucur2018improving}, influence and fairness \cite{gong2023influence,feng2023influence,razaghi2022group}, or influence and time/budget \cite{pham2019competitive,biswas2022multi}, either consider at most two, or occasionally \cite{biswas2022improved} three objectives, or model one of the objectives as an additional constraint. To the best of our knowledge, no work has addressed, so far, the problem of \emph{many-objective Influence Maximization}, i.e., solving the IM problem by explicitly handling more than three objectives at the same time.
\vspace{-.3cm}

\subsubsection*{\textbf{Contributions}}
In this work, we endeavor in a first attempt at effectively solving this many-objective IM problem. \Circled{1} We consider six different many-objective IM problem formulations, with up to six objectives simultaneously handled in terms of Pareto optimization. \Circled{2} We propose \algname (\algdesc), a Multi-Objective Evolutionary Algorithm (MOEA) based on NSGA-II \cite{996017} and leveraging on the graph-aware operators and smart initialization mechanism from \cite{IACCA2021100107}.
\Circled{3} We compare \algname in two experimental settings, including a total of nine graph datasets, two heuristic methods \cite{leskovec2007cost,goyal2011celf++}, another MOEA from \cite{bucur2018improving}, and a state-of-the-art DL technique, DeepIM \cite{ling2023deep}.
\Circled{4} Our experimental analysis reveals that in most cases \algname significantly outperforms the competitors in terms of achieved hypervolumes, proving to be a competitive solution for many-objective IM problems.
\Circled{5} We conduct a correlation analysis of the objective functions, to provide novel insights on IM.

The rest of the paper is structured as follows. In the next section, we introduce the main concepts on the IM problem. Then, \Cref{sec:rw} briefly summarizes the related work. The methods are presented in \Cref{sec:methods}. \Cref{sec:experiments} presents the experimental setup, while \Cref{sec:results} discusses the results. Finally, we draw the conclusions in \Cref{sec:conclusions}.


\section{Background}
\label{sec:background}
The IM problem is defined as a combinatorial optimization problem over a directed/undirected graph $\mathcal{G}$. The main goal of the problem is to find a subset of nodes, called \emph{seed set} ($S$), that can spread the influence the most under a given propagation model $\sigma$, i.e., $\argmax_{S} |\sigma(S)|$. The propagation model simulates the influence spread over discrete timesteps $t \in [0,\tau]$, such that at each timestep $t_i$ each of the \emph{active} nodes resulting from the previous timestep $t_{i-1}$ tries to activate (i.e., propagate the influence to) the nodes in its neighborhood. The pseudocode of the generic propagation model is provided in \Cref{alg:cascade}.

\begin{algorithm}[ht!]
	\caption{Propagation model $\sigma(S)$. $S$ is the seed set, $G$ is the graph, $\tau$ is the maximum number of timesteps, $p$ (needed only for IC) is the probability that an edge will be activated.} 
	\label{alg:cascade}
	{\small
		\begin{algorithmic}[1]
			\Ensure $S$
			\Require $\mathcal{G}$, $\tau$, [$p$]
			\State $A \leftarrow S$ \Comment{The set of activated nodes}
			\State $B \leftarrow S$ \Comment{Nodes activated in the previous timestep}
			\State $t=0$ \Comment{Timestep counter}
			\While {$B$ not empty \textbf{and} $t < \tau $}
				\State $C \leftarrow \emptyset$ \Comment{Nodes activated in the current timestep}
				\For {$n \in B$} 
					\For{$m \in \text{neighbours}(n) \setminus A$}
						\State $C \leftarrow C \cup \{m\}$ with probability $p$ \Comment{\Longunderstack[r]{Activation attempt\\($p$ differs for IC/WC/LT)}}\label{line:attempt}
					\EndFor
				\EndFor
				\State $B \leftarrow C$
				\State $A \leftarrow A \cup B$
				\State $t = t + 1$
			\EndWhile
			\State \Return $|A|$ \Comment{Total number of activated nodes}
		\end{algorithmic}
	}
\end{algorithm}

\begin{figure}[!ht]
 \centering
 \includegraphics[width=1\columnwidth]{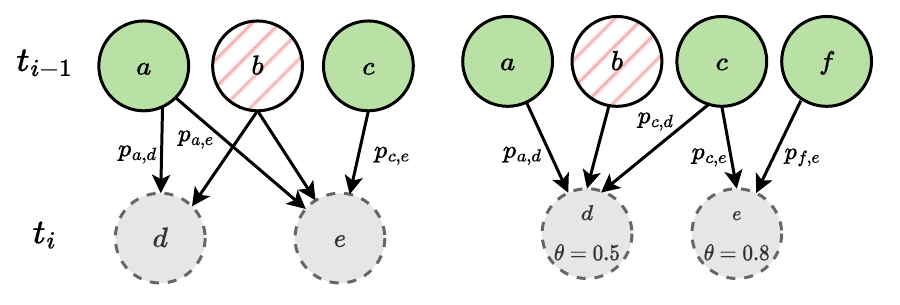}
 \caption{Graphical representation of the IC and WC models (left), and LT model (right). The top row shows activated nodes (green circles) and non-activated nodes (dashed red circles) at timestep $t_{i-1}$. The bottom row shows the nodes that can be activated at timestep $t_i$.}
 \label{fig:propagation}
 \vspace{-0.4cm}
\end{figure}

The activation attempt (line \ref{line:attempt} in the pseudocode) is probabilistic and depends on the specific propagation model considered.
While various models do exist, the most used 
are: the
\textbf{Independent Cascade} (IC), the \textbf{Weighted Cascade} (WC), and the \textbf{Linear Threshold} (LT) models, proposed in \cite{kempe2003maximizing}.
\Circled{1} In IC, the probability of activating a new node is based on a hyperparameter $p$, that is given as input to the propagation model.
Note that, here, $p$ is equal for all the nodes.
Referring to \Cref{fig:propagation} (left), $p_{i,j} = p~\forall~{i,j}$. 
\Circled{2} In WC, which can be seen as a non-parametric version of IC, the probability of activating a node $n$ is inversely proportional to its in-degree $d_{n}^{in}$ (i.e., $p_{m,n}=1/d_{n}^{in}$). Referring to \Cref{fig:propagation} (left), $p_{a,d} = 1/2$ and $p_{a,e} = p_{c,e} = 1/3$).
\Circled{3} Finally, in the LT model each edge has a weight $p$ computed as in the WC model, but the stochasticity relies on the threshold activation $\theta$ assigned to each node (the original model defined in \cite{kempe2003maximizing} sets uniformly samples $\theta$ in $[0,1]$ for each node), see \Cref{fig:propagation} (right). Namely, during the propagation process, a new node $n$ can be activated if the sum of its incoming weights (corresponding to the active nodes among its neighbours) is higher than its threshold $\theta_{n}$, i.e., $\sum_{m \in \text{neighbours}(n)} p_{m,n} \geq \theta_{n}$.

Due to the stochasticity of the propagation models, the influence propagation is computed through Monte Carlo simulations multiple times (usually $100$, see \cite{bucur2016influence}) for each seed set $S$, and then averaged.

\section{Related work}
\label{sec:rw}

We categorize the related works based on how they formulate the IM problem and, in particular, on the objectives they consider, which reflect the same objective functions we consider in our study.

\noindent \textbf{Influence}
For solving the IM problem, the majority of works builds upon the milestone work \cite{kempe2003maximizing}, which uses a greedy algorithm with approximation guarantees. To reduce the computational cost, several heuristics have been proposed \cite{leskovec2007cost,chen2009efficient,jiang2011simulated,wang2016maximizing}. A technique based on the statistical estimation of the number of nodes from which the seed set should be sampled has been proposed in \cite{tang2015influence}, and then extended to online processing \cite{tang2018online}.
As an alternative, various approaches based on EAs emerged \cite{bucur2016influence,lotf2022improved,chawla2023neighbor}, which have a computational advantage over the heuristics. The most recent line of works instead focuses on DL approaches, either based on RL \cite{li2022piano,chen2023touplegdd}, Learnable Embeddings \cite{panagopoulos2020multi,ling2023deep}, or combination of deep RL with EAs \cite{ma2022influence}.
Of note, all these works only focus on maximizing the influence, fixing a priori the seed set size.

\noindent \textbf{Seed set size}
Several works proposed to optimize, along with the influence, the seed set size, so to have a min-max IM problem formulation. The earliest works proposed the use of MOEAs \cite{bucur2017multi,bucur2018improving,cunegatti2022large}. In \cite{biswas2022improved}, the authors employ a MOEA to optimize seed set size, influence, and budget jointly. Other approaches employ swarm intelligence methods, such as in \cite{olivares2021multi,riquelme2023depth}, where the authors employ multi-objective Particle Swarm Optimization (with either scalarization or lexicographic ordering) to maximize the influence while minimizing the seed set size.

\noindent \textbf{Communities}\footnote{\emph{Communities} and \emph{fairness} have interchangeable definitions in the literature. In this section, we categorize them following the objective formulation defined in \Cref{sec:methods}.} This objective aims to fairly distribute the influenced nodes across all the parts (i.e., the communities) of the network.
In the literature, some works proposed to combine influence and communities into as single, monotone objective \cite{tsang2019group,ali2021fairness,rui2023scalable}, while other works focused on Pareto optimization \cite{gong2023influence,feng2023influence} or Graph Embeddings \cite{khajehnejad2020adversarial}.

\noindent \textbf{Fairness}
This objective aims to fairly allocate the nodes in the seed set w.r.t. the network communities they belong to. This objective has been defined in \cite{stoica2020seeding,becker2022fairness} and handled in various ways, e.g., through multi-objective constrained optimization \cite{tsang2019group,razaghi2022group}, integer programming \cite{farnad2020unifying}, and DL \cite{feng2023influence}.

\noindent \textbf{Budget}
The budget, along with the seed set size, is the most studied objective function in IM. The idea behind this objective is that each node in the seed set has a \emph{cost} (that is usually assumed to be proportional to its out-degree) that needs to be smaller than (or equal to) a fixed value.
While this problem could be trivially solved by selecting in the seed set the $k$ nodes with the highest out-degree in the graph, this may represent a sub-optimal solution, depending on the topological structure of the graph at hand.
Instead, it is more practical to consider only the sum of the out-degrees of the nodes in the seed set as the actual budget, without imposing any predefined limit on it.
Limited budget optimization has been explored in \cite{bian2020efficient,perrault2020budgeted,pham2019competitive}, while explicit multi-objective settings taking budget into account have been investigated in \cite{biswas2022multi}.

\noindent \textbf{Time}
As shown in \Cref{alg:cascade}, the propagation process can stop in two cases: either where no new nodes can be activated anymore, or once a given timestep limit $\tau$ is reached. Using as proxy for the time the number of timesteps of propagation is therefore crucial to reduce the computational cost needed to solve the IM problem.
The earliest work that attempted to optimize the propagation time over IM problem relies on a greedy approach \cite{liu2012time}. Later, specific algorithms have been proposed \cite{chen2012time,pham2019competitive} to avoid the complexity of greedy heuristics. The trade-off between influence and time has been empirically analyzed in \cite{tong2020time}, while the relation between fairness and time has been studied in \cite{ali2021fairness}.


\section{Methods}
\label{sec:methods}

We now introduce the objective functions adopted in the experimentation and the proposed \algname algorithm.

\subsection{Objective functions}
\label{subsec:obj}
We consider a total of six different objective functions\footnote{For clarity, we use the symbols ``$\uparrow$'' and ``$\downarrow$'' to indicate that an objective has to be maximized or minimized, respectively.} that are specific to the IM problem, namely:
\begin{enumerate}[leftmargin=*]
 \item \textbf{Influence} (\objI): This refers to the number of nodes in a graph that, starting from a given seed set $S$, are activated throughout the propagation process returned by \Cref{alg:cascade}. We refer to the set of nodes activated starting from the seed set $S$ as $\sigma(S)$. This function has to be \emph{maximized}.
 \item \textbf{Seed set size} (\objS): This is the number of nodes in the seed set ($|S|$), as given in input to \Cref{alg:cascade}. This function has to be \emph{minimized}.
 \item \textbf{Communities}: (\objC): For a given set of activated nodes that are not present in the seed set, namely $\sigma(S) \setminus S$, this metric measures the distance between the ideal distribution of activated nodes in each community (i.e., a uniform distribution), and the empirical distribution of activated nodes \cite{farnad2020unifying}. Throughout this whole work, the communities are referred to as subsets of nodes in a graph that are strongly connected among them while being less connected with the other parts of the graph, and have been computed using the Leiden algorithm \cite{traag2019louvain}.
 Since we treat Communities as a problem of matching an empirical distribution of probability with an ideal one, we employ the Jensen-Shannon distance, defined as:
 \begin{align}
 JSD(p, q) = \frac{1}{2} KL(p || M) + \frac{1}{2} KL(q || M)
 \end{align}
 where $KL$ is the Kullback-Leibler divergence, $p$ and $q$ are the two probability distribution functions, and $M=\frac{(p + q)}{2}$. The Jensen-Shannon divergence is bound in $[0, 1]$. As said, in the IM problem one of the probability distributions is fixed (namely, a uniform distribution $\mathcal{U}(1, c)$, where $c$ is the number of communities). Thus, we normalize $JSD$ by dividing it by the maximum $JSD$ distance that can occur in our setup, which is obtained by measuring the distance between a uniform distribution function and the Kronecker's delta function $\delta$ (which represents the case in which only nodes from a single community are influenced):
 \begin{align}
 JSD_N(p, \mathcal{U}(1, c)) = \frac{JSD(p, \mathcal{U}(1, c))}{JSD(\delta, \mathcal{U}(1, c))}
 \end{align}
 In other words, this metric measures if all communities are influenced fairly \emph{at the end} of the propagation process.
 This function has to be \emph{maximized}.
 \item \textbf{Fairness} (\objF): Fairness has the same core idea of Communities, i.e., optimizing the nodes' allocation throughout all the communities. However, differently from the previous metric that is calculated over $\sigma(S) \setminus S$, fairness is measured on the seed set $S$, i.e., it measures if the propagation \emph{starts} fairly from each community. It is computed exactly as Communities, but on $S$.
 This function has to be \emph{maximized}.
 \item \textbf{Budget} (\objB): The budget refers to the sum of the out-degrees of the nodes in the selected seed set $S$, i.e., $b=\sum_{n \in S}d_{n}^{out}$. This function has to be \emph{minimized}.
 \item \textbf{Time} (\objT): This is the total propagation time to influence the nodes for which we use, as a proxy, the number of timesteps in the \texttt{while} loop of \Cref{alg:cascade}. This function is usually referred to as the number of hops of the propagation model. This function has to be \emph{minimized}.
\end{enumerate}

\subsection{MOEIM}
\label{subsec:moeim}
Here we present \algname, which is composed of three main components, namely: (1) smart initialization (2) many-objective evolutionary optimization, and (3) graph-aware evolutionary operators, described below.

\subsubsection*{\textbf{Smart initialization}}
Each candidate solution represents a possible seed set (i.e., each solution is a set of nodes used as starting points for the propagation process). Practically, each individual consists in a set of integers where each integer is the id of a node used in the seed set. 
Taking inspiration from \cite{konotopska2021graph}, we apply a smart initialization technique that generates an initial population based on the individual capability of each node to influence the network.
More specifically, to improve diversity in the initial population, a ratio of $(1 - \lambda)$ initial solutions (with $\lambda$ being a hyperparameter) is initialized randomly with a size randomly sampled in $[1,k]$, while a ratio of $\lambda$ is created using the following procedure. First of all, we simulate the influence propagation of each \emph{single} node for $\tau=3$ hops. Then, we keep the best $k$ nodes (i.e., those that, taken individually, are able to influence the highest number of nodes in the graph), where $k$ is again the maximum size of the individual. Among these $k$ nodes, we filter out those with a out-degree lower than $\Theta$, which is another hyperparameter. Each individual in the $\lambda$ fraction of the initial population is then created with a size randomly sampled in $[1, k]$ by inserting the filtered nodes with probabilities proportional to their out-degree, akin to roulette selection.
\vspace{-0.2cm}

\subsubsection*{\textbf{Many-objective evolutionary optimization}}
For the evolutionary optimization core, we rely on the NSGA-II algorithm modified to solve the IM problem as in \cite{bucur2018improving}, and made available in \cite{IACCA2021100107}. Such an algorithm has been employed due to its ability to seamlessly add new objective functions to the optimization process.
However, unlike the existing multi-objective approaches to IM, which as seen earlier employ at most two or three objectives, here we extend the number of objectives up to six. In fact, we study: (1) the case in which we optimize 
\objI and \objS (as in \cite{bucur2017multi,bucur2018improving,cunegatti2022large}); (2) the case in which (\objI-\objS) plus each objective among \objC, \objF, \objB, and \objT are optimized, resulting in four three-objectives problems; and (3) the case in which we optimize all the metrics from \Cref{subsec:obj}, thus resulting in a six-objectives problem.
\vspace{-0.2cm}

\subsubsection*{\textbf{Graph-aware evolutionary operators}}
As demonstrated in previous works \cite{konotopska2021graph,chawla2023neighbor}, while standard one-point crossover is effective for obtaining state-of-the-art performance on the IM problem, graph-aware mutation operators can provide a dramatic performance boost w.r.t. na{\"i}ve mutations such as random node sampling.

Here, we employ a total of five different mutation operators previously introduced in \cite{konotopska2021graph}, of which two are stochastic, while three are graph-aware, to draw a balance between stochasticity and graph-based knowledge.
As for the stochastic operators, we use: (1) \emph{Global Random Insert Mutation} and (2) \emph{Global Random Removal Mutation}, which respectively randomly adds nodes to the individual, or removes nodes from it.
As for the graph-aware operators, we use: (1) \emph{Local Neighbor Mutation}, where a new node is randomly selected among the neighbors of the node to mutate; (2) \emph{Local Neighbor Second-Degree Mutation}, which selects a new node in the neighborhood of the nodes to mutate, with probability proportional to its out-degree; (3) \emph{Global Low Degree Mutation}, where the new node is selected from the whole graph with probability inversely proportional to its out-degree. 
At each mutation, one of the above mutation operators is selected with uniform probability.


\section{Experimental setup}
\label{sec:experiments}

We perform two separate experimental analyses of \algname. The first setting \setone compares our proposed method against two existing heuristics from \cite{wang2016maximizing,leskovec2007cost} and the state-of-the-art MOEA from \cite{bucur2018improving}. 
The second setting \settwo is focused on comparing the performances of \algname w.r.t. a state-of-the-art DL technique, DeepIM \cite{ling2023deep}.

\subsubsection*{\textbf{Datasets}}
For the first experimental setting \setone, we selected six different datasets to include directed and undirected, sparse and dense graphs (with an average out-degree from $6$ to $50$), ranging from $\sim$1k to $\sim$9k nodes and $\sim$25k to $\sim$100k edges. We also selected such datasets to cover a broad spectrum of values w.r.t. the number of communities, from $6$ to $46$. The selected datasets are: \textbf{email-eu-Core}$^{*}$ \cite{leskovec2007graph}, where nodes represent European research institutions and edges are emails exchanged among them; \textbf{facebook-combined}$^{*}$ \cite{leskovec2012learning}, where nodes are Facebook profiles and edges corresponds to friendship among users; \textbf{gnutella}$^{*}$ \cite{leskovec2007graph}, which is a peer-to-peer file sharing network; \textbf{wiki-vote}$^{*}$ \cite{leskovec2010signed}, where nodes corresponds to administrators of Wikipedia and edges corresponds to who-vote-for-whom during an internal election; \textbf{lastfm}$^{*}$ \cite{feather} is the social network LastFM of Asian users, where nodes are users and edges correspond to mutual following; and \textbf{Ca-HepTh}$^{*}$ \cite{leskovec2007graph} is a collaboration network of High Energy Physics Theory among the arXiv papers of who-cites-whom\footnote{The apex ``$^{*}$'' means that the dataset is available on the SNAP repository \cite{snapnets}, ``$^{+}$'' means that it is available on the Network Repository \cite{rossi2015network}, while ``$^{-}$'' refers to availability on the Netzschleuder network catalogue \cite{peixoto2020netzschleuder}}. 

For the second experimental setting \settwo, we selected the three publicly available datasets used in \cite{ling2023deep}, to allow for a direct comparison with the results presented in the original paper. In particular, the three datasets tested are \textbf{Jazz}$^{+,-}$ \cite{gleiser2003community}, which is a collaboration network among Jazz musicians/bands, \textbf{Cora-ML} \cite{mccallum2000automating}, where nodes are Computer Science papers and edges are citations among them, and \textbf{Power Grid}$^{+,-}$ \cite{watts1998collective}, where nodes are power relay points and edges correspond to line connecting them.

For both settings, each dataset has been pre-processed, to remove disconnected components (i.e., portions of the graph disconnected from the largest component of the graph) and low-quality communities. For that, we so relied on the pre-processing approach designed in \cite{rui2023scalable}, where each final graph represents the largest weakly connected components\footnote{In directed graphs, weakly (strongly) connected components are defined as the largest connected components of the graph whose nodes are reachable in at least one direction (both directions). In undirected graphs, weakly and strongly connected components coincide.} of the original graph and the communities with less than $10$ nodes are removed\footnote{To note that the final pre-processed graphs used in the experimental setting \settwo have the same properties as in the original paper \cite{ling2023deep}, given that the three datasets used in this setting do not present communities with less than 10 nodes.}.
The properties of all datasets, after this pre-processing, are available in \Cref{tab:dataset}.

\subsubsection*{\textbf{Baselines}}
For the first experimental setting \setone, we selected as baselines two heuristic algorithms, namely Generalized Degree Discount (GDD) \cite{wang2016maximizing} and Cost-Effective Lazy Forward selection (CELF) \cite{leskovec2007cost}. They are both designed to maximize the influence under a constraint on the seed set size, and they both work by adding nodes iteratively until reaching the selected budget.
More specifically, GDD adds nodes to the seed set nodes greedily based on the nodes' out-degree, while CELF is a faster version of the GDD designed to work as a hill-climbing two-pass greedy approach.
As a third baseline, we chose the MOEA presented in \cite{bucur2018improving}, since it was the first EA proposed for solving the multi-objective IM problem.

For the second experimental setting \settwo, we only compare to DeepIM \cite{ling2023deep}, since, from the numerical results reported in the original paper, this method outperformed other DL-based approaches when tested on the three selected datasets we tested in our experiments.

\begin{table}[ht!]
 \caption{Tested datasets. Top (bottom) rows: setting \setone (\settwo).}
 \label{tab:dataset}
 \centering
 \begin{adjustbox}{max width=1\columnwidth}
 \begin{tabular}{l rrr rrr rrr c}
 \toprule
 \multirow{2}[2]{*}{\textbf{Network}} & \multirow{2}[2]{*}{\textbf{Nodes}} & \multirow{2}[2]{*}{\textbf{Edges}} & \multicolumn{3}{c}{\textbf{Communities}} & \multicolumn{3}{c}{\textbf{Nodes' degree}} & \multirow{2}[2]{*}{\textbf{Type}} \\
 \cmidrule(lr){4-6} \cmidrule(lr){7-9}
 & & & \textbf{Num.} & \textbf{Min.} & \textbf{Max.} & \textbf{Avg.} & \textbf{Std.} & \textbf{Max.} \\
 \midrule
 email-eu-Core & 986 & 25552 & 7 & 57& 308& 51.83& 60.32& 546 & Directed \\
 facebook-combined & 4039 & 88234 & 17 & 19& 548& 43.69& 52.41& 1045 & Undirected\\
 gnutella & 6299 & 20776 & 20 & 11& 753& 6.6& 8.54& 97 & Directed \\ 
 wiki-vote & 7066 & 103663 & 6 & 29& 1959& 29.34& 60.55& 1167 & Directed \\ 
 lastfm & 7624 & 27806 & 23 & 11& 1015& 7.29& 11.5& 216 & Undirected \\ 
 CA-HepTh & 8638 & 24827 & 46 & 15& 625& 5.75& 6.46& 65 & Undirected\\ 
 \midrule
 Jazz & 198 & 2742 & 3 & 61& 73& 27.7& 17.41& 100 & Undirected\\ 
 Cora-ML & 2810 & 7981 & 22 & 12& 441& 5.68& 8.45& 246 & Undirected\\
 Power Grid & 4941 & 6594 & 38 & 31& 233& 2.67& 1.79& 19 & Undirected \\ 
 \bottomrule
 \end{tabular}
 \end{adjustbox}
 \vspace{-0.3cm}
\end{table}

\subsubsection*{\textbf{Hyperparameter setting}}

For the first experimental setting \setone, we set the maximum seed set size to $k=100$ for all the algorithms under comparison, as done in \cite{bucur2018improving}. Regarding the time limit ($\tau$), it is worth mentioning that several works \cite{lee2014fast,gong2016influence,konotopska2021graph} empirically showed how setting $\tau$ to a value as low as $2$ (i.e., 2-hop) provides a computationally efficient approximation of influence propagation, yielding final influence values that are practically equivalent to those achieved by CELF. Here, we opt for a value of $\tau=5$ (as \cite{gong2023influence}), which provides a fair compromise between the the 2-hop approximation and an unbounded spread process (i.e., $\tau=\infty$).
For a fair comparison, such time limit has been applied to all the baselines (GDD, CELF, and MOEA) as well as to \algname in the various combinations of objectives (see \Cref{sec:setting1}). 
In this experimental setting, we use two different propagation models (as in \citep{bucur2018improving,cunegatti2022large}), namely WC and IC, setting $p=0.05$ in the latter.

For the second experimental setting \settwo, for a fair comparison, we relied on the same IM setting adopted in \cite{ling2023deep}, where $k$ was set to $20\%$ of nodes in the graph, while no limits on $\tau$ and budget were applied. As in \cite{ling2023deep}, in this case we use as propagation models WC and LT, setting the threshold of each node uniformly sampled in $[0.3,0.6]$ in the latter.

For all the experiments, the evolutionary hyperparameters of \algname and MOEA \cite{bucur2018improving} have been kept fixed in both experimental settings \setone and \settwo, setting the population size to $100$, the number of offspring to $100$, the number of elites to $2$, the tournament size to $5$ (as in \cite{cunegatti2022large}), and the number of generations to $100$\footnote{Of note, in \cite{bucur2018improving} the number of generations was set to $500$. However, as we will show in the experimental results, \algname outperforms CELF in a much lower number of generations, keeping all the other parameters the same for both EAs. For this reason, and for the sake of computational efficiency, we reduced the number of generations to $100$ for both algorithms.}. For both experimental settings, the fitness evaluation of MOEA and \algname solutions have been computed by running 100 Monte Carlo simulations of the tested propagation model. 
Only for \algname the $\lambda=0.33$ of the initial population has been initiated with smart initialization presented in \Cref{subsec:moeim} where $\Theta$ is set to the average out-degree of the graph. The rest of the population is randomly initialized.

\subsection*{\textbf{Computational setup}}
To run the experiments, we employed a workstation with an Intel(R) Core(TM) i9-10980XE CPU $@3GHz$ with $18$ cores ($36$ threads) and 125 GB of RAM. Our code is implemented in Python3.8, running in multi-threading.


\begin{table*}[!ht]
 \caption{Avg. $\pm$ std. dev. hypervolume ($HV$) achieved by the algorithms under comparison and our proposed \algname, across 10 runs for \algname, MOEA, and GDD, and (because of computational limits, as discussed in \cite{bucur2018improving}) 3 runs for CELF (\objI: Influence; \objS: Seed set size; \objC: Communities; \objF: Fairness; \objB: Budget; \objT: Time).
 The column ${HV}_{\textrm{\objI-\objS}}$ indicates the $HV$ in the 2D (\objI-\objS) space, while the next four columns indicate the $HV$ calculated in the corresponding 3D space, and ${HV}_{\textrm{all}}$ indicates the $HV$ calculated in a 6D space in which all objectives (\objI-\objS-\objC-\objF-\objB-\objT) are considered. For \algname, we show the case where 2 objectives (\objI-\objS), the combinations of 3 objectives (\objI-\objS plus either \objC, \objF, \objB or \objS), and all objectives (\objI-\objS-\objC-\objF-\objB-\objT) are considered during the evolutionary process. The colored (underlined) cells indicate the highest (second highest) value per column, separately per each dataset, propagation model, and combination of objectives considered to calculate the $HV$.}
 \label{tab:results}
 \centering
 \begin{adjustbox}{max width=1\textwidth}
 \begin{tabular}{l l @{\hskip 0.1in} cccccc @{\hskip 0.3in} cccccc}
 
 \toprule
 \noalign{\bigskip}
 \multirow{2}[2]{*}{} & \multirow{2}[2]{*}{\textbf{Algorithm}} & \multicolumn{6}{c}{\textbf{Independent Cascade (IC)}} & \multicolumn{6}{c}{\textbf{Weighted Cascade (WC)}} \\
 \noalign{\smallskip}
 \cmidrule(lr{3em}){3-8} \cmidrule(lr{1em}){9-14}
 \noalign{\smallskip}
 & 
 & ${HV}_{\textrm{\objI-\objS}}$ & ${HV}_{\textrm{\objI-\objS-\objC}}$ & ${HV}_{\textrm{\objI-\objS-\objF}}$ & ${HV}_{\textrm{\objI-\objS-\objB}}$ & ${HV}_{\textrm{\objI-\objS-\objT}}$ & ${HV}_{\textrm{all}}$ & 
 ${HV}_{\textrm{\objI-\objS}}$ & ${HV}_{\textrm{\objI-\objS-\objC}}$ & ${HV}_{\textrm{\objI-\objS-\objF}}$ & ${HV}_{\textrm{\objI-\objS-\objB}}$ & ${HV}_{\textrm{\objI-\objS-\objT}}$ & ${HV}_{\textrm{all}}$ \\ 
 
 \noalign{\bigskip}
 \midrule
 \noalign{\bigskip}
 
 \multirow{9}[1]{*}{\STAB{\rotatebox[origin=c]{90}{email-eu-Core}}}
 & GDD \cite{wang2016maximizing} & $4.90{\textrm{e}}{-1} \pm3.72{\textrm{e}}{-5} $ &$4.62{\textrm{e}}{-1} \pm2.29{\textrm{e}}{-4} $ &$4.19{\textrm{e}}{-1} \pm3.82{\textrm{e}}{-6} $ &$4.57{\textrm{e}}{-1} \pm8.45{\textrm{e}}{-5} $ &$6.18{\textrm{e}}{-2} \pm3.62{\textrm{e}}{-3} $ &$3.21{\textrm{e}}{-2} \pm2.94{\textrm{e}}{-3}$ &\underline{$4.11{\textrm{e}}{-1} \pm2.26{\textrm{e}}{-4}$} &$3.78{\textrm{e}}{-1} \pm7.52{\textrm{e}}{-5} $ &$3.01{\textrm{e}}{-1} \pm1.21{\textrm{e}}{-4} $ & \cellcolor{green!80}$\mathbf{3.10{\textrm{e}}{-1} \pm4.47{\textrm{e}}{-4}}$ &$7.45{\textrm{e}}{-2} \pm5.62{\textrm{e}}{-3} $ &$2.68{\textrm{e}}{-2} \pm4.93{\textrm{e}}{-3}$\\
 & CELF \citep{leskovec2007cost} & $4.87{\textrm{e}}{-1} \pm8.97{\textrm{e}}{-4} $ &$4.58{\textrm{e}}{-1} \pm7.86{\textrm{e}}{-4} $ &$4.00{\textrm{e}}{-1} \pm3.54{\textrm{e}}{-3} $ &$4.60{\textrm{e}}{-1} \pm1.12{\textrm{e}}{-3} $ &$2.52{\textrm{e}}{-2} \pm1.43{\textrm{e}}{-2} $ &$1.43{\textrm{e}}{-2} \pm1.05{\textrm{e}}{-2}$ &$3.80{\textrm{e}}{-1} \pm1.40{\textrm{e}}{-3} $ &$3.47{\textrm{e}}{-1} \pm1.49{\textrm{e}}{-3} $ &$3.17{\textrm{e}}{-1} \pm6.99{\textrm{e}}{-3} $ &\underline{$3.07{\textrm{e}}{-1} \pm1.32{\textrm{e}}{-3}$} &$7.01{\textrm{e}}{-2} \pm1.62{\textrm{e}}{-3} $ &$3.59{\textrm{e}}{-2} \pm3.62{\textrm{e}}{-3}$\\
 & MOEA \cite{bucur2018improving} & $4.96{\textrm{e}}{-1} \pm1.84{\textrm{e}}{-3} $ &$4.68{\textrm{e}}{-1} \pm1.80{\textrm{e}}{-3} $ &$4.17{\textrm{e}}{-1} \pm1.39{\textrm{e}}{-2} $ &$4.53{\textrm{e}}{-1} \pm2.14{\textrm{e}}{-3} $ &$1.97{\textrm{e}}{-2} \pm2.79{\textrm{e}}{-2} $ &$3.00{\textrm{e}}{-3} \pm9.49{\textrm{e}}{-3}$ &$3.59{\textrm{e}}{-1} \pm6.48{\textrm{e}}{-3} $ &$3.26{\textrm{e}}{-1} \pm6.57{\textrm{e}}{-3} $ &$3.15{\textrm{e}}{-1} \pm7.68{\textrm{e}}{-3} $ &$2.53{\textrm{e}}{-1} \pm1.84{\textrm{e}}{-3} $ &$4.97{\textrm{e}}{-2} \pm5.22{\textrm{e}}{-3} $ &$2.11{\textrm{e}}{-2} \pm5.25{\textrm{e}}{-3}$ \\
 
 \cmidrule(lr{3em}){3-8} \cmidrule(lr{1em}){9-14}
 & MOEIM (\objI-\objS) & \cellcolor{blue!70}$\mathbf{4.99{\textrm{e}}{-1} \pm5.99{\textrm{e}}{-4}}$ & \underline{$4.70{\textrm{e}}{-1} \pm5.59{\textrm{e}}{-4}$} &$4.18{\textrm{e}}{-1} \pm4.33{\textrm{e}}{-3} $ &$4.54{\textrm{e}}{-1} \pm3.18{\textrm{e}}{-3} $ &$0.00{\textrm{e}}{0} \pm0.00{\textrm{e}}{0}$ &$3.43{\textrm{e}}{-4} \pm1.08{\textrm{e}}{-3}$ &\cellcolor{blue!70}$\mathbf{4.13{\textrm{e}}{-1} \pm1.37{\textrm{e}}{-3}}$ &\cellcolor{green!80}$\mathbf{3.82{\textrm{e}}{-1} \pm1.30{\textrm{e}}{-3}}$ &$3.52{\textrm{e}}{-1} \pm1.63{\textrm{e}}{-2} $ &$2.56{\textrm{e}}{-1} \pm3.49{\textrm{e}}{-4} $ &$7.06{\textrm{e}}{-2} \pm5.46{\textrm{e}}{-3} $ &$1.67{\textrm{e}}{-2} \pm6.13{\textrm{e}}{-3}$\\
 & MOEIM (\objI-\objS-\objC) & $4.95{\textrm{e}}{-1} \pm1.68{\textrm{e}}{-3} $ & \cellcolor{green!80}$\mathbf{4.72{\textrm{e}}{-1} \pm1.53{\textrm{e}}{-3}}$ &$4.17{\textrm{e}}{-1} \pm5.05{\textrm{e}}{-3} $ &$4.54{\textrm{e}}{-1} \pm1.17{\textrm{e}}{-3} $ &$1.92{\textrm{e}}{-2} \pm2.49{\textrm{e}}{-2} $ &$3.08{\textrm{e}}{-3} \pm4.51{\textrm{e}}{-3}$ &$4.09{\textrm{e}}{-1} \pm1.20{\textrm{e}}{-3} $ &\underline{$3.80{\textrm{e}}{-1} \pm1.20{\textrm{e}}{-3} $} &\underline{$3.76{\textrm{e}}{-1} \pm2.09{\textrm{e}}{-3}$} &$2.57{\textrm{e}}{-1} \pm8.29{\textrm{e}}{-4} $ &$7.98{\textrm{e}}{-2} \pm4.57{\textrm{e}}{-3} $ &$2.76{\textrm{e}}{-2} \pm4.99{\textrm{e}}{-3}$\\
 
 & MOEIM (\objI-\objS-\objF) & $4.95{\textrm{e}}{-1} \pm6.50{\textrm{e}}{-4} $ &$4.67{\textrm{e}}{-1} \pm8.70{\textrm{e}}{-4} $ & \cellcolor{green!80}$\mathbf{4.64{\textrm{e}}{-1} \pm1.48{\textrm{e}}{-3}}$ &$4.55{\textrm{e}}{-1} \pm2.05{\textrm{e}}{-3} $ &$5.67{\textrm{e}}{-3} \pm4.11{\textrm{e}}{-3} $ &$2.31{\textrm{e}}{-3} \pm5.26{\textrm{e}}{-3}$ &$4.06{\textrm{e}}{-1} \pm1.39{\textrm{e}}{-3} $ &$2.11{\textrm{e}}{-1} \pm1.98{\textrm{e}}{-3} $ &\cellcolor{green!80}$\mathbf{3.89{\textrm{e}}{-1} \pm1.28{\textrm{e}}{-3}}$ &$2.60{\textrm{e}}{-1} \pm1.79{\textrm{e}}{-3} $ &$7.61{\textrm{e}}{-2} \pm2.28{\textrm{e}}{-3} $ &$3.40{\textrm{e}}{-2} \pm4.03{\textrm{e}}{-3}$\\
 & MOEIM (\objI-\objS-\objB) & $4.91{\textrm{e}}{-1} \pm1.82{\textrm{e}}{-3} $ &$4.63{\textrm{e}}{-1} \pm1.83{\textrm{e}}{-3} $ &$4.41{\textrm{e}}{-1} \pm4.35{\textrm{e}}{-3} $ & \cellcolor{green!80}$\mathbf{4.74{\textrm{e}}{-1} \pm1.70{\textrm{e}}{-3}}$ &$1.38{\textrm{e}}{-1} \pm2.81{\textrm{e}}{-3} $ &$3.45{\textrm{e}}{-2} \pm6.53{\textrm{e}}{-3}$ &$3.94{\textrm{e}}{-1} \pm9.16{\textrm{e}}{-4} $ &$3.63{\textrm{e}}{-1} \pm6.83{\textrm{e}}{-4} $ &$3.61{\textrm{e}}{-1} \pm2.82{\textrm{e}}{-3} $ &$2.66{\textrm{e}}{-1} \pm1.16{\textrm{e}}{-3}$ &$8.83{\textrm{e}}{-2} \pm6.41{\textrm{e}}{-4} $ &\underline{$3.97{\textrm{e}}{-2} \pm1.45{\textrm{e}}{-3}$} \\
 
 & MOEIM (\objI-\objS-\objT) & \underline{$4.98{\textrm{e}}{-1} \pm6.78{\textrm{e}}{-4} $}&$4.69{\textrm{e}}{-1} \pm5.90{\textrm{e}}{-4} $ &$4.22{\textrm{e}}{-1} \pm8.15{\textrm{e}}{-3} $ &\underline{$4.64{\textrm{e}}{-1} \pm3.29{\textrm{e}}{-4}$} &\underline{$1.60{\textrm{e}}{-1} \pm5.66{\textrm{e}}{-3}$} &\underline{$4.37{\textrm{e}}{-2} \pm,3.57{\textrm{e}}{-3}$} &$4.10{\textrm{e}}{-1} \pm7.90{\textrm{e}}{-4} $ &$3.79{\textrm{e}}{-1} \pm7.93{\textrm{e}}{-4} $ &$3.61{\textrm{e}}{-1} \pm5.18{\textrm{e}}{-3} $ &$2.59{\textrm{e}}{-1} \pm1.23{\textrm{e}}{-3} $ &\cellcolor{green!80}$\mathbf{9.44{\textrm{e}}{-2} \pm7.78{\textrm{e}}{-4}}$ &$3.56{\textrm{e}}{-2} \pm8.13{\textrm{e}}{-4}$ \\
 
 & MOEIM (all) & $4.86{\textrm{e}}{-1} \pm5.43{\textrm{e}}{-4} $ &$4.61{\textrm{e}}{-1} \pm6.62{\textrm{e}}{-4} $ &\underline{$4.46{\textrm{e}}{-1} \pm2.55{\textrm{e}}{-3}$} &$4.66{\textrm{e}}{-1} \pm8.19{\textrm{e}}{-4} $ & \cellcolor{green!80}$\mathbf{1.60{\textrm{e}}{-1} \pm1.30{\textrm{e}}{-3}}$ & $\cellcolor{red!70}\mathbf{7.29{\textrm{e}}{-2} \pm3.77{\textrm{e}}{-3}}$ &$3.81{\textrm{e}}{-1} \pm2.42{\textrm{e}}{-3} $ &$3.50{\textrm{e}}{-1} \pm2.35{\textrm{e}}{-3} $ &$3.57{\textrm{e}}{-1} \pm1.36{\textrm{e}}{-3} $ &$2.55{\textrm{e}}{-1} \pm8.23{\textrm{e}}{-4} $ &\underline{$8.99{\textrm{e}}{-2} \pm4.20{\textrm{e}}{-4}$} &$\cellcolor{red!70}\mathbf{4.25{\textrm{e}}{-2} \pm7.43{\textrm{e}}{-4}}$ \\

 \noalign{\bigskip}
 \midrule
 \noalign{\bigskip}
 
 \multirow{9}[1]{*}{\STAB{\rotatebox[origin=c]{90}{facebook-combined}}}
 & GDD \cite{wang2016maximizing} & $5.15{\textrm{e}}{-1} \pm4.22{\textrm{e}}{-5} $ &$3.65{\textrm{e}}{-1} \pm4.30{\textrm{e}}{-4} $ &$2.95{\textrm{e}}{-1} \pm2.38{\textrm{e}}{-5} $ &$4.62{\textrm{e}}{-1} \pm2.92{\textrm{e}}{-4} $ &$0.00{\textrm{e}}{0} \pm0.00{\textrm{e}}{0}$ &$0.00{\textrm{e}}{0} \pm0.00{\textrm{e}}{0}$ &$2.07{\textrm{e}}{-1} \pm9.42{\textrm{e}}{-5} $ &$1.34{\textrm{e}}{-1} \pm4.89{\textrm{e}}{-5} $ &$8.76{\textrm{e}}{-2} \pm3.38{\textrm{e}}{-5} $ &$1.70{\textrm{e}}{-1} \pm5.91{\textrm{e}}{-5} $ &$4.66{\textrm{e}}{-3} \pm3.30{\textrm{e}}{-3} $ &$1.81{\textrm{e}}{-4} \pm1.91{\textrm{e}}{-4}$\\
 
 & CELF \citep{leskovec2007cost} & $5.18{\textrm{e}}{-1} \pm4.99{\textrm{e}}{-4} $ &$3.65{\textrm{e}}{-1} \pm1.08{\textrm{e}}{-3} $ &$2.84{\textrm{e}}{-1} \pm9.19{\textrm{e}}{-3} $ &$4.73{\textrm{e}}{-1} \pm1.64{\textrm{e}}{-3} $ &$0.00{\textrm{e}}{0} \pm0.00{\textrm{e}}{0}$ &$0.00{\textrm{e}}{0} \pm0.00{\textrm{e}}{0}$ &$2.46{\textrm{e}}{-1} \pm1.10{\textrm{e}}{-3} $ &$1.84{\textrm{e}}{-1} \pm7.23{\textrm{e}}{-4} $ &$1.37{\textrm{e}}{-1} \pm3.06{\textrm{e}}{-3} $ &$2.04{\textrm{e}}{-1} \pm6.49{\textrm{e}}{-4} $ &$2.33{\textrm{e}}{-3} \pm3.29{\textrm{e}}{-3} $ &$1.18{\textrm{e}}{-4} \pm2.05{\textrm{e}}{-4}$\\
 & MOEA \cite{bucur2018improving} & $5.02{\textrm{e}}{-1} \pm4.35{\textrm{e}}{-3} $ &$3.47{\textrm{e}}{-1} \pm5.72{\textrm{e}}{-3} $ &$2.98{\textrm{e}}{-1} \pm5.97{\textrm{e}}{-3} $ &$4.74{\textrm{e}}{-1} \pm1.52{\textrm{e}}{-3} $ &$0.00{\textrm{e}}{0} \pm0.00{\textrm{e}}{0}$ &$0.00{\textrm{e}}{0} \pm0.00{\textrm{e}}{0}$ &$1.96{\textrm{e}}{-1} \pm5.32{\textrm{e}}{-3} $ &$1.32{\textrm{e}}{-1} \pm7.79{\textrm{e}}{-3} $ &$1.21{\textrm{e}}{-1} \pm6.25{\textrm{e}}{-3} $ &$1.71{\textrm{e}}{-1} \pm3.28{\textrm{e}}{-3} $ &$0.00{\textrm{e}}{0} \pm0.00{\textrm{e}}{0}$ &$1.25{\textrm{e}}{-4} \pm1.49{\textrm{e}}{-4}$ \\
 \cmidrule(lr{3em}){3-8} \cmidrule(lr{1em}){9-14}
 & MOEIM (\objI-\objS) & \cellcolor{blue!70} $\mathbf{5.22{\textrm{e}}{-1} \pm8.53{\textrm{e}}{-4}}$ &$\underline{3.78{\textrm{e}}{-1} \pm5.26{\textrm{e}}{-3}}$ &$3.18{\textrm{e}}{-1} \pm5.60{\textrm{e}}{-3} $ &$4.80{\textrm{e}}{-1} \pm4.06{\textrm{e}}{-3} $ &$0.00{\textrm{e}}{0} \pm0.00{\textrm{e}}{0}$ &$0.00{\textrm{e}}{0} \pm0.00{\textrm{e}}{0}$ &\cellcolor{blue!70}$\mathbf{2.50{\textrm{e}}{-1} \pm2.78{\textrm{e}}{-4}}$ &\underline{$1.89{\textrm{e}}{-1} \pm9.27{\textrm{e}}{-4}$} &$1.49{\textrm{e}}{-1} \pm9.40{\textrm{e}}{-4} $ &$2.06{\textrm{e}}{-1} \pm1.25{\textrm{e}}{-3} $ &$8.34{\textrm{e}}{-3} \pm3.15{\textrm{e}}{-3} $ &$2.56{\textrm{e}}{-4} \pm2.42{\textrm{e}}{-4}$\\
 & MOEIM (\objI-\objS-\objC) & $5.19{\textrm{e}}{-1} \pm1.03{\textrm{e}}{-3} $ &\cellcolor{green!80}$\mathbf{3.93{\textrm{e}}{-1} \pm1.88{\textrm{e}}{-3}}$ &$3.10{\textrm{e}}{-1} \pm9.67{\textrm{e}}{-3} $ &$4.78{\textrm{e}}{-1} \pm1.31{\textrm{e}}{-3} $ &$0.00{\textrm{e}}{0} \pm0.00{\textrm{e}}{0}$ &$1.58{\textrm{e}}{-4} \pm5.01{\textrm{e}}{-4}$ &$2.46{\textrm{e}}{-1} \pm1.15{\textrm{e}}{-3} $ &\cellcolor{green!80}$\mathbf{1.95{\textrm{e}}{-1} \pm8.70{\textrm{e}}{-4}}$ &$1.65{\textrm{e}}{-1} \pm3.67{\textrm{e}}{-3} $ &$2.04{\textrm{e}}{-1} \pm1.37{\textrm{e}}{-3} $ &$4.09{\textrm{e}}{-3} \pm2.98{\textrm{e}}{-3} $ &$1.97{\textrm{e}}{-4} \pm2.19{\textrm{e}}{-4}$ \\
 & MOEIM (\objI-\objS-\objF) & $5.14{\textrm{e}}{-1} \pm1.38{\textrm{e}}{-3} $ &$3.71{\textrm{e}}{-1} \pm4.36{\textrm{e}}{-4} $ &\cellcolor{green!80} $\mathbf{3.73{\textrm{e}}{-1} \pm2.11{\textrm{e}}{-3}}$ &$\underline{4.86{\textrm{e}}{-1} \pm3.54{\textrm{e}}{-3}}$ &$3.18{\textrm{e}}{-3} \pm4.49{\textrm{e}}{-3} $ &$1.13{\textrm{e}}{-4} \pm2.39{\textrm{e}}{-4}$ &$2.41{\textrm{e}}{-1} \pm1.75{\textrm{e}}{-3} $ &$1.87{\textrm{e}}{-1} \pm2.13{\textrm{e}}{-3} $ &\cellcolor{green!80}$\mathbf{1.79{\textrm{e}}{-1} \pm1.40{\textrm{e}}{-3}}$ &$2.02{\textrm{e}}{-1} \pm1.22{\textrm{e}}{-3} $ &$8.46{\textrm{e}}{-3} \pm2.61{\textrm{e}}{-3} $ &$6.99{\textrm{e}}{-4} \pm6.83{\textrm{e}}{-4}$\\
 & MOEIM (\objI-\objS-\objB) & $5.03{\textrm{e}}{-1} \pm1.40{\textrm{e}}{-3} $ &$3.54{\textrm{e}}{-1} \pm2.86{\textrm{e}}{-3} $ &$3.28{\textrm{e}}{-1} \pm6.31{\textrm{e}}{-4} $ &\cellcolor{green!80}$\mathbf{4.91{\textrm{e}}{-1} \pm1.28{\textrm{e}}{-3}}$ &\cellcolor{green!80}$\mathbf{7.66{\textrm{e}}{-2} \pm4.09{\textrm{e}}{-3}}$ &\underline{$6.72{\textrm{e}}{-3} \pm1.52{\textrm{e}}{-3}$ }&$2.41{\textrm{e}}{-1} \pm2.13{\textrm{e}}{-3} $ &$1.80{\textrm{e}}{-1} \pm2.15{\textrm{e}}{-3} $ &$1.48{\textrm{e}}{-1} \pm2.27{\textrm{e}}{-3} $ &\cellcolor{green!80}$\mathbf{2.07{\textrm{e}}{-1} \pm1.47{\textrm{e}}{-3}}$ & \underline{$1.70{\textrm{e}}{-2} \pm1.75{\textrm{e}}{-3} $} &$2.12{\textrm{e}}{-3} \pm6.82{\textrm{e}}{-4}$\\
 & MOEIM (\objI-\objS-\objT) & \underline{$5.21{\textrm{e}}{-1} \pm3.96{\textrm{e}}{-4}$} &$3.70{\textrm{e}}{-1} \pm3.58{\textrm{e}}{-3} $ &$3.11{\textrm{e}}{-1} \pm1.25{\textrm{e}}{-2} $ &$4.82{\textrm{e}}{-1} \pm5.86{\textrm{e}}{-4} $ &\underline{$7.59{\textrm{e}}{-2} \pm2.84{\textrm{e}}{-3}$} &$5.86{\textrm{e}}{-3} \pm1.25{\textrm{e}}{-3}$ &\underline{$2.48{\textrm{e}}{-1} \pm7.09{\textrm{e}}{-4}$} &$1.88{\textrm{e}}{-1} \pm1.06{\textrm{e}}{-3} $ &$1.51{\textrm{e}}{-1} \pm2.52{\textrm{e}}{-3} $ &\underline{$2.06{\textrm{e}}{-1} \pm4.93{\textrm{e}}{-4}$} &\cellcolor{green!80}$\mathbf{1.85{\textrm{e}}{-2} \pm1.55{\textrm{e}}{-3}}$ &\underline{$2.20{\textrm{e}}{-3} \pm3.36{\textrm{e}}{-4}$} \\
 & MOEIM (all) & $4.91{\textrm{e}}{-1} \pm1.66{\textrm{e}}{-3} $ &$3.53{\textrm{e}}{-1} \pm3.71{\textrm{e}}{-3} $ &$\underline{3.50{\textrm{e}}{-1} \pm1.44{\textrm{e}}{-3}}$ &$4.78{\textrm{e}}{-1} \pm1.31{\textrm{e}}{-3} $ &$6.40{\textrm{e}}{-2} \pm2.02{\textrm{e}}{-3} $ &$\cellcolor{red!70}\mathbf{9.22{\textrm{e}}{-3} \pm1.01{\textrm{e}}{-3}}$ &$2.22{\textrm{e}}{-1} \pm4.23{\textrm{e}}{-3} $ &$1.66{\textrm{e}}{-1} \pm3.35{\textrm{e}}{-3} $ &$1.60{\textrm{e}}{-1} \pm2.19{\textrm{e}}{-3} $ &$1.88{\textrm{e}}{-1} \pm3.00{\textrm{e}}{-3} $ &$1.61{\textrm{e}}{-2} \pm3.21{\textrm{e}}{-4} $ &$\cellcolor{red!70}\mathbf{3.89{\textrm{e}}{-3} \pm3.53{\textrm{e}}{-4}}$ \\

 \noalign{\bigskip}
 \midrule
 \noalign{\bigskip}
 
 \multirow{9}[1]{*}{\STAB{\rotatebox[origin=c]{90}{gnutella}}}
 & GDD \cite{wang2016maximizing} & $\cellcolor{blue!70}1.55{\textrm{e}}{-2} \pm7.03{\textrm{e}}{-6} $ &$\cellcolor{green!80}\mathbf{1.01{\textrm{e}}{-2} \pm9.04{\textrm{e}}{-6}}$ &$9.90{\textrm{e}}{-3} \pm3.79{\textrm{e}}{-6} $ &$9.06{\textrm{e}}{-3} \pm7.26{\textrm{e}}{-6} $ &$9.75{\textrm{e}}{-3} \pm1.46{\textrm{e}}{-4} $ &$1.84{\textrm{e}}{-3} \pm1.63{\textrm{e}}{-5}$ &$1.64{\textrm{e}}{-1} \pm1.64{\textrm{e}}{-4} $ &$1.43{\textrm{e}}{-1} \pm1.30{\textrm{e}}{-4} $ &$1.04{\textrm{e}}{-1} \pm1.07{\textrm{e}}{-4} $ &$1.09{\textrm{e}}{-1} \pm1.51{\textrm{e}}{-4} $ &$6.95{\textrm{e}}{-3} \pm2.25{\textrm{e}}{-3} $ &$1.23{\textrm{e}}{-3} \pm6.39{\textrm{e}}{-4}$\\
 & CELF \citep{leskovec2007cost} & $1.53{\textrm{e}}{-2} \pm1.11{\textrm{e}}{-4} $ &$8.34{\textrm{e}}{-3} \pm1.71{\textrm{e}}{-4} $ &$7.81{\textrm{e}}{-3} \pm4.94{\textrm{e}}{-4} $ &$9.48{\textrm{e}}{-3} \pm3.19{\textrm{e}}{-5} $ &$8.58{\textrm{e}}{-3} \pm1.65{\textrm{e}}{-4} $ &$1.33{\textrm{e}}{-3} \pm1.56{\textrm{e}}{-4}$ &\cellcolor{blue!70}$\mathbf{2.93{\textrm{e}}{-1} \pm2.29{\textrm{e}}{-3}}$ &\cellcolor{green!80}$\mathbf{2.68{\textrm{e}}{-1} \pm2.68{\textrm{e}}{-3}}$ &$1.93{\textrm{e}}{-1} \pm4.30{\textrm{e}}{-3} $ &\cellcolor{green!80}$\mathbf{2.30{\textrm{e}}{-1} \pm1.69{\textrm{e}}{-3}} $ &$3.63{\textrm{e}}{-3} \pm3.76{\textrm{e}}{-5} $ &$3.67{\textrm{e}}{-4} \pm1.43{\textrm{e}}{-6}$\\
 & MOEA \cite{bucur2018improving} & $1.34{\textrm{e}}{-2} \pm1.55{\textrm{e}}{-4} $ &$7.48{\textrm{e}}{-3} \pm2.36{\textrm{e}}{-4} $ &$9.33{\textrm{e}}{-3} \pm9.90{\textrm{e}}{-5} $ &$8.58{\textrm{e}}{-3} \pm6.39{\textrm{e}}{-5} $ &$8.51{\textrm{e}}{-3} \pm8.23{\textrm{e}}{-5} $ &$1.70{\textrm{e}}{-3} \pm7.55{\textrm{e}}{-5}$ &$1.86{\textrm{e}}{-1} \pm5.41{\textrm{e}}{-3} $ &$1.64{\textrm{e}}{-1} \pm5.68{\textrm{e}}{-3} $ &$1.22{\textrm{e}}{-1} \pm4.75{\textrm{e}}{-3} $ &$1.57{\textrm{e}}{-1} \pm5.12{\textrm{e}}{-3} $ &$1.93{\textrm{e}}{-3} \pm1.39{\textrm{e}}{-3} $ &$1.92{\textrm{e}}{-4} \pm2.06{\textrm{e}}{-4}$ \\
 \cmidrule(lr{3em}){3-8} \cmidrule(lr{1em}){9-14}
 & MOEIM (\objI-\objS) & \underline{$1.55{\textrm{e}}{-2} \pm1.22{\textrm{e}}{-4} $} &$9.54{\textrm{e}}{-3} \pm1.43{\textrm{e}}{-4} $ &$1.06{\textrm{e}}{-2} \pm2.45{\textrm{e}}{-4} $ &$8.34{\textrm{e}}{-3} \pm3.02{\textrm{e}}{-5} $ &$9.06{\textrm{e}}{-3} \pm1.38{\textrm{e}}{-4} $ &$1.61{\textrm{e}}{-3} \pm7.57{\textrm{e}}{-5}$ &\underline{$2.90{\textrm{e}}{-1} \pm1.26{\textrm{e}}{-3} $} &\underline{$2.66{\textrm{e}}{-1} \pm9.10{\textrm{e}}{-4}$} &$2.08{\textrm{e}}{-1} \pm1.24{\textrm{e}}{-3} $ &$2.06{\textrm{e}}{-1} \pm2.21{\textrm{e}}{-3} $ &$2.83{\textrm{e}}{-3} \pm6.74{\textrm{e}}{-4} $ &$2.22{\textrm{e}}{-4} \pm1.38{\textrm{e}}{-4}$\\
 & MOEIM (\objI-\objS-\objC) & $1.52{\textrm{e}}{-2} \pm9.41{\textrm{e}}{-5} $ &\underline{$9.73{\textrm{e}}{-3} \pm1.18{\textrm{e}}{-4}$} &\underline{$1.18{\textrm{e}}{-2} \pm2.29{\textrm{e}}{-4}$} &$8.28{\textrm{e}}{-3} \pm6.27{\textrm{e}}{-5} $ &$9.50{\textrm{e}}{-3} \pm2.05{\textrm{e}}{-4} $ &$2.02{\textrm{e}}{-3} \pm1.06{\textrm{e}}{-4}$ &$2.80{\textrm{e}}{-1} \pm3.58{\textrm{e}}{-3} $ &$2.61{\textrm{e}}{-1} \pm3.37{\textrm{e}}{-3} $ &$2.07{\textrm{e}}{-1} \pm6.51{\textrm{e}}{-3} $ &$2.00{\textrm{e}}{-1} \pm3.09{\textrm{e}}{-3} $ &$0.00{\textrm{e}}{0} \pm0.00{\textrm{e}}{0}$ &$3.46{\textrm{e}}{-4} \pm2.95{\textrm{e}}{-4}$ \\
 & MOEIM (\objI-\objS-\objF) & $1.43{\textrm{e}}{-2} \pm7.34{\textrm{e}}{-5} $ &$8.75{\textrm{e}}{-3} \pm8.07{\textrm{e}}{-5} $ &$\cellcolor{green!80}\mathbf{1.23{\textrm{e}}{-2} \pm1.55{\textrm{e}}{-4}}$ &$8.57{\textrm{e}}{-3} \pm3.50{\textrm{e}}{-4} $ &$9.37{\textrm{e}}{-3} \pm9.72{\textrm{e}}{-5} $ &$2.47{\textrm{e}}{-3} \pm1.16{\textrm{e}}{-4}$ &$2.71{\textrm{e}}{-1} \pm1.58{\textrm{e}}{-3} $ &$2.49{\textrm{e}}{-1} \pm1.53{\textrm{e}}{-3} $ &\cellcolor{green!80}$\mathbf{2.33{\textrm{e}}{-1} \pm4.13{\textrm{e}}{-3}}$ &$1.90{\textrm{e}}{-1} \pm1.83{\textrm{e}}{-3} $ &$6.40{\textrm{e}}{-3} \pm2.02{\textrm{e}}{-3} $ &$8.61{\textrm{e}}{-4} \pm3.24{\textrm{e}}{-4}$\\
 & MOEIM (\objI-\objS-\objB) & $1.43{\textrm{e}}{-2} \pm8.04{\textrm{e}}{-5} $ &$8.52{\textrm{e}}{-3} \pm1.50{\textrm{e}}{-4} $ &$1.11{\textrm{e}}{-2} \pm2.64{\textrm{e}}{-4} $ &$\cellcolor{green!80}\mathbf{1.05{\textrm{e}}{-2} \pm6.82{\textrm{e}}{-5}}$ &$1.04{\textrm{e}}{-2} \pm3.95{\textrm{e}}{-4} $ &$2.75{\textrm{e}}{-3} \pm1.08{\textrm{e}}{-4}$ &$2.82{\textrm{e}}{-1} \pm1.56{\textrm{e}}{-3} $ &$2.59{\textrm{e}}{-1} \pm8.24{\textrm{e}}{-4} $ &\underline{$2.12{\textrm{e}}{-1} \pm1.37{\textrm{e}}{-3}$} &\underline{$2.19{\textrm{e}}{-1} \pm2.68{\textrm{e}}{-3}$} &$3.98{\textrm{e}}{-3} \pm7.49{\textrm{e}}{-4} $ &$4.47{\textrm{e}}{-4} \pm1.46{\textrm{e}}{-4}$\\
 & MOEIM (\objI-\objS-\objT) & $1.52{\textrm{e}}{-2} \pm2.64{\textrm{e}}{-4} $ &$9.27{\textrm{e}}{-3} \pm3.50{\textrm{e}}{-4} $ &$1.08{\textrm{e}}{-2} \pm4.58{\textrm{e}}{-4} $ &$9.27{\textrm{e}}{-3} \pm9.49{\textrm{e}}{-5} $ &$\cellcolor{green!80}\mathbf{1.14{\textrm{e}}{-2} \pm1.94{\textrm{e}}{-4}}$ &$2.26{\textrm{e}}{-3} \pm1.19{\textrm{e}}{-4}$ &$2.81{\textrm{e}}{-1} \pm2.54{\textrm{e}}{-3} $ &$2.57{\textrm{e}}{-1} \pm2.44{\textrm{e}}{-3} $ &$2.01{\textrm{e}}{-1} \pm4.29{\textrm{e}}{-3} $ &$1.98{\textrm{e}}{-1} \pm2.72{\textrm{e}}{-3} $ &\cellcolor{green!80}$\mathbf{1.74{\textrm{e}}{-2} \pm1.26{\textrm{e}}{-3}} $ &$4.28{\textrm{e}}{-3} \pm8.34{\textrm{e}}{-4}$ \\
 & MOEIM (all) & $1.36{\textrm{e}}{-2} \pm4.89{\textrm{e}}{-5} $ &$7.86{\textrm{e}}{-3} \pm8.30{\textrm{e}}{-5} $ &$1.11{\textrm{e}}{-2} \pm1.81{\textrm{e}}{-4} $ &\underline{$1.01{\textrm{e}}{-2} \pm2.77{\textrm{e}}{-5}$} &\underline{$1.05{\textrm{e}}{-2} \pm1.65{\textrm{e}}{-4}$} &$\cellcolor{red!70}\mathbf{2.81{\textrm{e}}{-3} \pm5.96{\textrm{e}}{-5}}$ &$2.46{\textrm{e}}{-1} \pm1.92{\textrm{e}}{-3} $ &$2.26{\textrm{e}}{-1} \pm2.33{\textrm{e}}{-3} $ &$2.03{\textrm{e}}{-1} \pm7.72{\textrm{e}}{-4} $ &$1.73{\textrm{e}}{-1} \pm1.25{\textrm{e}}{-3} $ &\underline{$1.57{\textrm{e}}{-2} \pm1.40{\textrm{e}}{-3}$} &$\cellcolor{red!70}\mathbf{5.53{\textrm{e}}{-3} \pm3.62{\textrm{e}}{-4}}$ \\
 
 \noalign{\bigskip}
 \midrule
 \noalign{\bigskip}
 
 \multirow{9}[1]{*}{\STAB{\rotatebox[origin=c]{90}{wiki-vote}}}
 & GDD \cite{wang2016maximizing} & $1.79{\textrm{e}}{-1} \pm3.33{\textrm{e}}{-5} $ &$1.70{\textrm{e}}{-1} \pm4.82{\textrm{e}}{-5} $ &$1.63{\textrm{e}}{-1} \pm3.54{\textrm{e}}{-5} $ &$1.59{\textrm{e}}{-1} \pm7.96{\textrm{e}}{-5} $ &$2.13{\textrm{e}}{-2} \pm2.28{\textrm{e}}{-3} $ &$7.07{\textrm{e}}{-3} \pm1.62{\textrm{e}}{-3}$ &$6.94{\textrm{e}}{-2} \pm1.15{\textrm{e}}{-5} $ &$6.60{\textrm{e}}{-2} \pm2.61{\textrm{e}}{-5} $ &$6.43{\textrm{e}}{-2} \pm9.02{\textrm{e}}{-6} $ &$5.10{\textrm{e}}{-2} \pm2.64{\textrm{e}}{-5} $ &$1.47{\textrm{e}}{-2} \pm6.13{\textrm{e}}{-4} $ &$8.78{\textrm{e}}{-3} \pm4.62{\textrm{e}}{-5}$\\
 & CELF \citep{leskovec2007cost} & $1.76{\textrm{e}}{-1} \pm1.01{\textrm{e}}{-3} $ &$1.67{\textrm{e}}{-1} \pm9.63{\textrm{e}}{-4} $ &$1.35{\textrm{e}}{-1} \pm9.46{\textrm{e}}{-3} $ &$1.63{\textrm{e}}{-1} \pm4.87{\textrm{e}}{-4} $ &$5.96{\textrm{e}}{-3} \pm8.43{\textrm{e}}{-3} $ &$8.49{\textrm{e}}{-4} \pm1.47{\textrm{e}}{-3}$ &$8.06{\textrm{e}}{-2} \pm8.86{\textrm{e}}{-4} $ &$7.48{\textrm{e}}{-2} \pm9.77{\textrm{e}}{-4} $ &$6.88{\textrm{e}}{-2} \pm2.10{\textrm{e}}{-3} $ &$\cellcolor{green!80} \mathbf{5.96{\textrm{e}}{-2} \pm4.37{\textrm{e}}{-4} }$ &$1.66{\textrm{e}}{-2} \pm7.10{\textrm{e}}{-4} $ &$9.01{\textrm{e}}{-3} \pm2.32{\textrm{e}}{-4}$\\
 & MOEA \cite{bucur2018improving} & $1.67{\textrm{e}}{-1} \pm1.62{\textrm{e}}{-3} $ &$1.58{\textrm{e}}{-1} \pm1.06{\textrm{e}}{-3} $ &$1.55{\textrm{e}}{-1} \pm1.55{\textrm{e}}{-3} $ &$1.57{\textrm{e}}{-1} \pm1.34{\textrm{e}}{-3} $ &$0.00{\textrm{e}}{0} \pm0.00{\textrm{e}}{0}$ &$0.00{\textrm{e}}{0} \pm0.00{\textrm{e}}{0}$ &$5.20{\textrm{e}}{-2} \pm1.49{\textrm{e}}{-3} $ &$4.69{\textrm{e}}{-2} \pm1.75{\textrm{e}}{-3} $ &$4.59{\textrm{e}}{-2} \pm2.30{\textrm{e}}{-3} $ &$4.33{\textrm{e}}{-2} \pm1.15{\textrm{e}}{-3} $ &$1.13{\textrm{e}}{-2} \pm9.34{\textrm{e}}{-4} $ &$6.96{\textrm{e}}{-3} \pm4.38{\textrm{e}}{-4}$ \\
 \cmidrule(lr{3em}){3-8} \cmidrule(lr{1em}){9-14}
 & MOEIM (\objI-\objS) & \cellcolor{blue!70}$\mathbf{1.84{\textrm{e}}{-1} \pm1.24{\textrm{e}}{-4}}$ &$\cellcolor{green!80}\mathbf{1.75{\textrm{e}}{-1} \pm1.96{\textrm{e}}{-4}}$ &$1.45{\textrm{e}}{-1} \pm2.03{\textrm{e}}{-3} $ &$1.68{\textrm{e}}{-1} \pm6.91{\textrm{e}}{-4} $ &$1.66{\textrm{e}}{-2} \pm4.14{\textrm{e}}{-3} $ &$4.69{\textrm{e}}{-3} \pm2.62{\textrm{e}}{-3}$ &\cellcolor{blue!70}$\mathbf{8.53{\textrm{e}}{-2} \pm6.97{\textrm{e}}{-5}}$ &$\cellcolor{green!80}\mathbf{8.02{\textrm{e}}{-2} \pm1.02{\textrm{e}}{-4} }$ &$7.82{\textrm{e}}{-2} \pm1.11{\textrm{e}}{-3} $ &$5.60{\textrm{e}}{-2} \pm7.76{\textrm{e}}{-5} $ &$1.80{\textrm{e}}{-2} \pm2.54{\textrm{e}}{-4} $ &$9.25{\textrm{e}}{-3} \pm1.51{\textrm{e}}{-4}$\\
 & MOEIM (\objI-\objS-\objC) & $1.81{\textrm{e}}{-1} \pm4.71{\textrm{e}}{-4} $ &\underline{$1.74{\textrm{e}}{-1} \pm4.62{\textrm{e}}{-4} $}&$1.58{\textrm{e}}{-1} \pm5.29{\textrm{e}}{-3} $ &$1.67{\textrm{e}}{-1} \pm6.26{\textrm{e}}{-4} $ &$1.70{\textrm{e}}{-2} \pm8.58{\textrm{e}}{-4} $ &$5.10{\textrm{e}}{-3} \pm1.16{\textrm{e}}{-3}$ &$8.28{\textrm{e}}{-2} \pm1.97{\textrm{e}}{-4} $ &$7.92{\textrm{e}}{-2} \pm2.76{\textrm{e}}{-4} $ &\underline{$8.08{\textrm{e}}{-2} \pm2.49{\textrm{e}}{-4}$} &$5.57{\textrm{e}}{-2} \pm2.95{\textrm{e}}{-4} $ &$1.77{\textrm{e}}{-2} \pm1.23{\textrm{e}}{-4} $ &$1.02{\textrm{e}}{-2} \pm2.30{\textrm{e}}{-4}$ \\
 & MOEIM (\objI-\objS-\objF) & $1.81{\textrm{e}}{-1} \pm4.93{\textrm{e}}{-4} $ &$1.72{\textrm{e}}{-1} \pm4.26{\textrm{e}}{-4} $ &$\cellcolor{green!80}\mathbf{1.78{\textrm{e}}{-1} \pm4.11{\textrm{e}}{-4}}$ &$1.67{\textrm{e}}{-1} \pm6.52{\textrm{e}}{-4} $ &$1.11{\textrm{e}}{-2} \pm1.34{\textrm{e}}{-3} $ &$4.77{\textrm{e}}{-3} \pm1.29{\textrm{e}}{-3}$&$8.29{\textrm{e}}{-2} \pm2.60{\textrm{e}}{-4} $ &$7.89{\textrm{e}}{-2} \pm2.71{\textrm{e}}{-4} $ &$\cellcolor{green!80}\mathbf{8.20{\textrm{e}}{-2} \pm1.91{\textrm{e}}{-4}}$ &$5.64{\textrm{e}}{-2} \pm2.81{\textrm{e}}{-4} $ &$1.80{\textrm{e}}{-2} \pm1.53{\textrm{e}}{-4} $ &$1.06{\textrm{e}}{-2} \pm1.25{\textrm{e}}{-4}$\\
 & MOEIM (\objI-\objS-\objB) & $1.80{\textrm{e}}{-1} \pm7.94{\textrm{e}}{-4} $ &$1.71{\textrm{e}}{-1} \pm9.19{\textrm{e}}{-4} $ & \underline{$1.68{\textrm{e}}{-1} \pm6.98{\textrm{e}}{-4}$} &$\cellcolor{green!80}\mathbf{1.69{\textrm{e}}{-1} \pm8.32{\textrm{e}}{-4}}$ &$2.18{\textrm{e}}{-2} \pm2.84{\textrm{e}}{-3} $ &$1.26{\textrm{e}}{-2} \pm1.94{\textrm{e}}{-4}$ &$7.94{\textrm{e}}{-2} \pm4.30{\textrm{e}}{-4} $ &$7.47{\textrm{e}}{-2} \pm4.30{\textrm{e}}{-4} $ &$7.55{\textrm{e}}{-2} \pm6.72{\textrm{e}}{-4} $ &\underline{$5.74{\textrm{e}}{-2} \pm1.11{\textrm{e}}{-4}$} &$2.07{\textrm{e}}{-2} \pm2.80{\textrm{e}}{-4} $ & $1.21{\textrm{e}}{-2} \pm2.44{\textrm{e}}{-3}$ \\
 & MOEIM (\objI-\objS-\objT) & \underline{$1.83{\textrm{e}}{-1} \pm1.03{\textrm{e}}{-3} $} &$1.74{\textrm{e}}{-1} \pm9.38{\textrm{e}}{-4} $ &$1.48{\textrm{e}}{-1} \pm3.14{\textrm{e}}{-3} $ & \underline{$1.69{\textrm{e}}{-1} \pm5.45{\textrm{e}}{-4}$} &$\cellcolor{green!80}\mathbf{3.99{\textrm{e}}{-2} \pm4.09{\textrm{e}}{-4}} $ &\underline{$1.50{\textrm{e}}{-2} \pm2.21{\textrm{e}}{-3}$} &\underline{$8.41{\textrm{e}}{-2} \pm1.52{\textrm{e}}{-4}$}&\underline{$7.92{\textrm{e}}{-2} \pm1.62{\textrm{e}}{-4}$} &$7.71{\textrm{e}}{-2} \pm1.03{\textrm{e}}{-3} $ &$5.62{\textrm{e}}{-2} \pm1.33{\textrm{e}}{-4} $ &$\cellcolor{green!80}\mathbf{2.34{\textrm{e}}{-2} \pm4.99{\textrm{e}}{-5}}$ & \underline{$1.30{\textrm{e}}{-2} \pm3.65{\textrm{e}}{-4}$} \\
 & MOEIM (all) & $1.72{\textrm{e}}{-1} \pm4.71{\textrm{e}}{-4} $ &$1.64{\textrm{e}}{-1} \pm4.69{\textrm{e}}{-4} $ &$1.67{\textrm{e}}{-1} \pm7.59{\textrm{e}}{-4} $ &$1.59{\textrm{e}}{-1} \pm2.45{\textrm{e}}{-4} $ &\underline{$3.69{\textrm{e}}{-2} \pm9.70{\textrm{e}}{-4}$} &$\cellcolor{red!70}\mathbf{2.04{\textrm{e}}{-2} \pm6.60{\textrm{e}}{-4}}$ &$7.55{\textrm{e}}{-2} \pm7.02{\textrm{e}}{-4} $ &$7.16{\textrm{e}}{-2} \pm6.84{\textrm{e}}{-4} $ &$7.32{\textrm{e}}{-2} \pm6.19{\textrm{e}}{-4} $ &$5.29{\textrm{e}}{-2} \pm5.15{\textrm{e}}{-4} $ &\underline{$2.15{\textrm{e}}{-2} \pm1.85{\textrm{e}}{-4}$} &$\cellcolor{red!70}\mathbf{1.32{\textrm{e}}{-2} \pm1.75{\textrm{e}}{-4}}$ \\
 
 \noalign{\bigskip}
 \midrule
 \noalign{\bigskip}
 
 \multirow{9}[1]{*}{\STAB{\rotatebox[origin=c]{90}{lastfm}}}
 & GDD \cite{wang2016maximizing} & \cellcolor{blue!70}$\mathbf{7.16{\textrm{e}}{-2} \pm4.14{\textrm{e}}{-5}}$ &$\cellcolor{green!80}\mathbf{4.02{\textrm{e}}{-2} \pm3.49{\textrm{e}}{-5} }$ &$3.67{\textrm{e}}{-2} \pm2.45{\textrm{e}}{-5} $ &$5.45{\textrm{e}}{-2} \pm1.81{\textrm{e}}{-5} $ &$3.03{\textrm{e}}{-3} \pm7.27{\textrm{e}}{-4} $ &$2.25{\textrm{e}}{-4} \pm1.67{\textrm{e}}{-4}$ &$1.88{\textrm{e}}{-1} \pm1.37{\textrm{e}}{-4} $ &$1.20{\textrm{e}}{-1} \pm1.25{\textrm{e}}{-4} $ &$7.52{\textrm{e}}{-2} \pm6.00{\textrm{e}}{-5} $ &$1.33{\textrm{e}}{-1} \pm7.10{\textrm{e}}{-5} $ &$3.75{\textrm{e}}{-3} \pm1.06{\textrm{e}}{-3} $ &$2.89{\textrm{e}}{-4} \pm6.84{\textrm{e}}{-5}$\\
 & CELF \citep{leskovec2007cost} & $6.31{\textrm{e}}{-2} \pm6.68{\textrm{e}}{-4} $ &$3.15{\textrm{e}}{-2} \pm4.13{\textrm{e}}{-4} $ &$2.48{\textrm{e}}{-2} \pm6.73{\textrm{e}}{-4} $ &$5.35{\textrm{e}}{-2} \pm1.58{\textrm{e}}{-4} $ &$2.63{\textrm{e}}{-3} \pm4.80{\textrm{e}}{-4} $ &$1.27{\textrm{e}}{-4} \pm6.40{\textrm{e}}{-5}$ &$1.90{\textrm{e}}{-1} \pm1.70{\textrm{e}}{-3} $ &$1.27{\textrm{e}}{-1} \pm6.78{\textrm{e}}{-4} $ &$9.32{\textrm{e}}{-2} \pm3.35{\textrm{e}}{-4} $ &$1.41{\textrm{e}}{-1} \pm1.25{\textrm{e}}{-3} $ &$5.66{\textrm{e}}{-3} \pm9.10{\textrm{e}}{-4} $ &$4.16{\textrm{e}}{-4} \pm1.43{\textrm{e}}{-3}$\\
 & MOEA \cite{bucur2018improving} & $5.35{\textrm{e}}{-2} \pm6.52{\textrm{e}}{-5} $ &$2.66{\textrm{e}}{-2} \pm4.43{\textrm{e}}{-4} $ &$2.68{\textrm{e}}{-2} \pm1.30{\textrm{e}}{-3} $ &$4.78{\textrm{e}}{-2} \pm5.21{\textrm{e}}{-5} $ &$2.04{\textrm{e}}{-3} \pm8.84{\textrm{e}}{-4} $ &$1.67{\textrm{e}}{-4} \pm7.88{\textrm{e}}{-5}$ &$1.09{\textrm{e}}{-1} \pm4.73{\textrm{e}}{-3} $ &$6.77{\textrm{e}}{-2} \pm2.67{\textrm{e}}{-3} $ &$5.62{\textrm{e}}{-2} \pm3.37{\textrm{e}}{-3} $ &$9.44{\textrm{e}}{-2} \pm3.43{\textrm{e}}{-3} $ &$4.79{\textrm{e}}{-3} \pm1.67{\textrm{e}}{-3} $ &$6.51{\textrm{e}}{-4} \pm4.08{\textrm{e}}{-4}$ \\
 \cmidrule(lr{3em}){3-8} \cmidrule(lr{1em}){9-14}
 & MOEIM (\objI-\objS) &\underline{ $6.76{\textrm{e}}{-2} \pm5.83{\textrm{e}}{-4} $ }&$3.55{\textrm{e}}{-2} \pm4.56{\textrm{e}}{-4} $ &$3.11{\textrm{e}}{-2} \pm5.66{\textrm{e}}{-5} $ &$5.45{\textrm{e}}{-2} \pm3.51{\textrm{e}}{-4} $ &$1.94{\textrm{e}}{-3} \pm2.60{\textrm{e}}{-4} $ &$9.18{\textrm{e}}{-5} \pm4.95{\textrm{e}}{-5}$ &\cellcolor{blue!70}$\mathbf{1.97{\textrm{e}}{-1} \pm5.55{\textrm{e}}{-4}}$ &$1.37{\textrm{e}}{-1} \pm6.39{\textrm{e}}{-4} $ &$1.07{\textrm{e}}{-1} \pm2.05{\textrm{e}}{-3} $ &$\cellcolor{green!80}\mathbf{1.44{\textrm{e}}{-1} \pm3.74{\textrm{e}}{-4}}$ &$5.09{\textrm{e}}{-3} \pm1.26{\textrm{e}}{-3} $ &$3.49{\textrm{e}}{-4} \pm2.03{\textrm{e}}{-4}$\\
 & MOEIM (\objI-\objS-\objC) & $6.67{\textrm{e}}{-2} \pm4.76{\textrm{e}}{-4} $ &\underline{$3.71{\textrm{e}}{-2} \pm2.08{\textrm{e}}{-4} $} &$3.60{\textrm{e}}{-2} \pm7.13{\textrm{e}}{-4} $ &\underline{$5.47{\textrm{e}}{-2} \pm2.59{\textrm{e}}{-4} $}&$3.34{\textrm{e}}{-3} \pm5.70{\textrm{e}}{-4} $ &$3.12{\textrm{e}}{-4} \pm1.95{\textrm{e}}{-4}$ &$1.89{\textrm{e}}{-1} \pm7.93{\textrm{e}}{-4} $ &$\cellcolor{green!80}\mathbf{1.42{\textrm{e}}{-1} \pm1.72{\textrm{e}}{-3}}$ &\underline{$1.20{\textrm{e}}{-1} \pm2.96{\textrm{e}}{-3} $} &$1.42{\textrm{e}}{-1} \pm4.56{\textrm{e}}{-4} $ &$6.64{\textrm{e}}{-3} \pm1.67{\textrm{e}}{-3} $ &$8.15{\textrm{e}}{-4} \pm3.02{\textrm{e}}{-4}$ \\
 & MOEIM (\objI-\objS-\objF) & $6.38{\textrm{e}}{-2} \pm3.91{\textrm{e}}{-4} $ &$3.44{\textrm{e}}{-2} \pm6.78{\textrm{e}}{-4} $ &$\cellcolor{green!80}\mathbf{4.28{\textrm{e}}{-2} \pm3.51{\textrm{e}}{-4}}$ &$5.43{\textrm{e}}{-2} \pm4.21{\textrm{e}}{-4} $ &$4.09{\textrm{e}}{-3} \pm1.14{\textrm{e}}{-3} $ &$7.51{\textrm{e}}{-4} \pm2.17{\textrm{e}}{-4}$ &$1.83{\textrm{e}}{-1} \pm7.87{\textrm{e}}{-4} $ &$1.30{\textrm{e}}{-1} \pm1.79{\textrm{e}}{-3} $ &$\cellcolor{green!80}\mathbf{1.26{\textrm{e}}{-1} \pm8.53{\textrm{e}}{-4}}$ &$1.39{\textrm{e}}{-1} \pm6.79{\textrm{e}}{-4} $ &$7.03{\textrm{e}}{-3} \pm1.05{\textrm{e}}{-3} $ &$1.07{\textrm{e}}{-3} \pm3.15{\textrm{e}}{-4}$\\
 & MOEIM (\objI-\objS-\objB) & $6.33{\textrm{e}}{-2} \pm5.11{\textrm{e}}{-4} $ &$3.29{\textrm{e}}{-2} \pm2.77{\textrm{e}}{-4} $ &$3.57{\textrm{e}}{-2} \pm1.13{\textrm{e}}{-3} $ &$\cellcolor{green!80}\mathbf{5.54{\textrm{e}}{-2} \pm3.63{\textrm{e}}{-4}}$ &$7.62{\textrm{e}}{-3} \pm3.53{\textrm{e}}{-4} $ &$1.09{\textrm{e}}{-3} \pm1.37{\textrm{e}}{-4}$ &$1.81{\textrm{e}}{-1} \pm1.40{\textrm{e}}{-3} $ &$1.26{\textrm{e}}{-1} \pm7.01{\textrm{e}}{-4} $ &$1.07{\textrm{e}}{-1} \pm4.39{\textrm{e}}{-4} $ &$1.40{\textrm{e}}{-1} \pm9.64{\textrm{e}}{-4} $ &$9.97{\textrm{e}}{-3} \pm5.51{\textrm{e}}{-4} $ & \underline{$2.21{\textrm{e}}{-3} \pm2.89{\textrm{e}}{-4}$} \\
 & MOEIM (\objI-\objS-\objT) & $6.67{\textrm{e}}{-2} \pm6.05{\textrm{e}}{-4} $ &$3.47{\textrm{e}}{-2} \pm8.44{\textrm{e}}{-4} $ &$3.13{\textrm{e}}{-2} \pm8.41{\textrm{e}}{-4} $ &$5.42{\textrm{e}}{-2} \pm2.81{\textrm{e}}{-4} $ &$\cellcolor{green!80}\mathbf{9.79{\textrm{e}}{-3} \pm5.70{\textrm{e}}{-4}}$ &\underline{$1.37{\textrm{e}}{-3} \pm1.55{\textrm{e}}{-4}$} &\underline{$1.93{\textrm{e}}{-1} \pm2.48{\textrm{e}}{-3}$} &\underline{$1.32{\textrm{e}}{-1} \pm2.99{\textrm{e}}{-3}$} &$1.02{\textrm{e}}{-1} \pm4.55{\textrm{e}}{-3} $ &\underline{$1.42{\textrm{e}}{-1} \pm5.50{\textrm{e}}{-4}$} &\underline{$1.22{\textrm{e}}{-2} \pm7.10{\textrm{e}}{-4}$} &$1.69{\textrm{e}}{-3} \pm2.37{\textrm{e}}{-4}$ \\
 & MOEIM (all) & $6.15{\textrm{e}}{-2} \pm2.95{\textrm{e}}{-4} $ &$3.26{\textrm{e}}{-2} \pm2.63{\textrm{e}}{-4} $ &\underline{$3.90{\textrm{e}}{-2} \pm1.64{\textrm{e}}{-4} $}&$5.40{\textrm{e}}{-2} \pm2.45{\textrm{e}}{-4} $ &\underline{$8.96{\textrm{e}}{-3} \pm1.57{\textrm{e}}{-4}$} &$\cellcolor{red!70}\mathbf{1.76{\textrm{e}}{-3} \pm8.95{\textrm{e}}{-5}}$ &$1.69{\textrm{e}}{-1} \pm5.81{\textrm{e}}{-4} $ &$1.18{\textrm{e}}{-1} \pm5.37{\textrm{e}}{-4} $ &$1.09{\textrm{e}}{-1} \pm1.08{\textrm{e}}{-3} $ &$1.29{\textrm{e}}{-1} \pm4.31{\textrm{e}}{-4} $ &$\cellcolor{green!80}\mathbf{1.27{\textrm{e}}{-2} \pm3.73{\textrm{e}}{-4}}$ &$\cellcolor{red!70}\mathbf{3.14{\textrm{e}}{-3} \pm3.21{\textrm{e}}{-4}}$ \\
 \noalign{\bigskip}
 \midrule
 \noalign{\bigskip}
 
 \multirow{9}[1]{*}{\STAB{\rotatebox[origin=c]{90}{CA-HepTh}}}
 & GDD \cite{wang2016maximizing} &\cellcolor{blue!70} $\mathbf{2.90{\textrm{e}}{-2} \pm1.44{\textrm{e}}{-5}}$ &$\cellcolor{green!80}\mathbf{1.43{\textrm{e}}{-2} \pm1.72{\textrm{e}}{-5}}$ &\underline{$1.41{\textrm{e}}{-2} \pm5.81{\textrm{e}}{-6}$} &\underline{$2.08{\textrm{e}}{-2} \pm1.94{\textrm{e}}{-5}$} &$6.31{\textrm{e}}{-3} \pm2.13{\textrm{e}}{-5} $ &$8.43{\textrm{e}}{-4} \pm8.78{\textrm{e}}{-6}$ &$8.75{\textrm{e}}{-2} \pm3.64{\textrm{e}}{-5} $ &$5.26{\textrm{e}}{-2} \pm4.03{\textrm{e}}{-5} $ &$3.34{\textrm{e}}{-2} \pm1.32{\textrm{e}}{-5} $ &$5.99{\textrm{e}}{-2} \pm2.81{\textrm{e}}{-5} $ &$2.52{\textrm{e}}{-3} \pm3.48{\textrm{e}}{-4} $ &$3.79{\textrm{e}}{-4} \pm1.80{\textrm{e}}{-4}$\\
 & CELF \citep{leskovec2007cost} & $2.62{\textrm{e}}{-2} \pm2.80{\textrm{e}}{-4} $ &$1.09{\textrm{e}}{-2} \pm1.62{\textrm{e}}{-4} $ &$8.86{\textrm{e}}{-3} \pm1.56{\textrm{e}}{-4} $ &$2.00{\textrm{e}}{-2} \pm1.17{\textrm{e}}{-4} $ &$5.47{\textrm{e}}{-3} \pm5.46{\textrm{e}}{-5} $ &$4.99{\textrm{e}}{-4} \pm 1.17{\textrm{e}}{-5}$ &$9.17{\textrm{e}}{-2} \pm1.07{\textrm{e}}{-4} $ &$5.64{\textrm{e}}{-2} \pm3.88{\textrm{e}}{-4} $ &$4.21{\textrm{e}}{-2} \pm1.35{\textrm{e}}{-3} $ &$6.61{\textrm{e}}{-2} \pm2.87{\textrm{e}}{-4} $ &$4.72{\textrm{e}}{-3} \pm1.07{\textrm{e}}{-3} $ &$5.17{\textrm{e}}{-4} \pm2.31{\textrm{e}}{-4}$\\
 & MOEA \cite{bucur2018improving} & $2.00{\textrm{e}}{-2} \pm6.46{\textrm{e}}{-4} $ &$7.84{\textrm{e}}{-3} \pm4.22{\textrm{e}}{-4} $ &$9.63{\textrm{e}}{-3} \pm4.26{\textrm{e}}{-4} $ &$1.73{\textrm{e}}{-2} \pm3.95{\textrm{e}}{-4} $ &$4.24{\textrm{e}}{-3} \pm1.34{\textrm{e}}{-4} $ &$5.84{\textrm{e}}{-4} \pm3.28{\textrm{e}}{-5}$ &$5.69{\textrm{e}}{-2} \pm1.15{\textrm{e}}{-3} $ &$3.35{\textrm{e}}{-2} \pm1.11{\textrm{e}}{-3} $ &$2.87{\textrm{e}}{-2} \pm5.24{\textrm{e}}{-4} $ &$4.84{\textrm{e}}{-2} \pm9.00{\textrm{e}}{-4} $ &$4.70{\textrm{e}}{-3} \pm8.12{\textrm{e}}{-4} $ &$5.90{\textrm{e}}{-4} \pm2.18{\textrm{e}}{-4}$ \\
 \cmidrule(lr{3em}){3-8} \cmidrule(lr{1em}){9-14}
 & MOEIM (\objI-\objS) & \underline{$2.76{\textrm{e}}{-2} \pm1.10{\textrm{e}}{-4}$} &$1.23{\textrm{e}}{-2} \pm7.88{\textrm{e}}{-5} $ &$1.12{\textrm{e}}{-2} \pm1.76{\textrm{e}}{-4} $ &$2.07{\textrm{e}}{-2} \pm8.60{\textrm{e}}{-5} $ &$5.66{\textrm{e}}{-3} \pm7.38{\textrm{e}}{-5} $ &$6.40{\textrm{e}}{-4} \pm1.81{\textrm{e}}{-5}$ &\cellcolor{blue!70}$9.43{\textrm{e}}{-2} \pm3.42{\textrm{e}}{-4} $ &$6.01{\textrm{e}}{-2} \pm2.19{\textrm{e}}{-4} $ &$4.84{\textrm{e}}{-2} \pm1.72{\textrm{e}}{-3} $ &\underline{$6.80{\textrm{e}}{-2} \pm2.39{\textrm{e}}{-4} $ }&$5.40{\textrm{e}}{-3} \pm5.99{\textrm{e}}{-4} $ &$5.84{\textrm{e}}{-4} \pm1.22{\textrm{e}}{-4}$\\
 & MOEIM (\objI-\objS-\objC) & $2.62{\textrm{e}}{-2} \pm1.77{\textrm{e}}{-4} $ &\underline{$1.30{\textrm{e}}{-2} \pm1.06{\textrm{e}}{-4}$} &$1.28{\textrm{e}}{-2} \pm3.01{\textrm{e}}{-4} $ &$2.02{\textrm{e}}{-2} \pm8.37{\textrm{e}}{-5} $ &$5.61{\textrm{e}}{-3} \pm1.10{\textrm{e}}{-4} $ &$8.74{\textrm{e}}{-4} \pm3.34{\textrm{e}}{-5}$ &$9.20{\textrm{e}}{-2} \pm4.02{\textrm{e}}{-4} $ &$\cellcolor{green!80}\mathbf{6.23{\textrm{e}}{-2} \pm9.97{\textrm{e}}{-5}}$ & \underline{$5.29{\textrm{e}}{-2} \pm2.16{\textrm{e}}{-4}$} &$6.80{\textrm{e}}{-2} \pm3.06{\textrm{e}}{-4} $ &$5.02{\textrm{e}}{-3} \pm6.67{\textrm{e}}{-4} $ &$7.26{\textrm{e}}{-4} \pm2.06{\textrm{e}}{-4}$ \\
 & MOEIM (\objI-\objS-\objF) & $2.48{\textrm{e}}{-2} \pm2.22{\textrm{e}}{-4} $ &$1.11{\textrm{e}}{-2} \pm1.03{\textrm{e}}{-4} $ &$\cellcolor{green!80}\mathbf{1.48{\textrm{e}}{-2} \pm1.45{\textrm{e}}{-4}}$ &$2.04{\textrm{e}}{-2} \pm9.91{\textrm{e}}{-5} $ &$6.42{\textrm{e}}{-3} \pm8.81{\textrm{e}}{-5} $ &\underline{$1.23{\textrm{e}}{-3} \pm4.80{\textrm{e}}{-5}$ }&$8.76{\textrm{e}}{-2} \pm8.54{\textrm{e}}{-4} $ &$5.73{\textrm{e}}{-2} \pm9.90{\textrm{e}}{-4} $ &$\cellcolor{green!80}\mathbf{5.50{\textrm{e}}{-2} \pm1.32{\textrm{e}}{-3}}$ &$6.63{\textrm{e}}{-2} \pm2.58{\textrm{e}}{-4} $ &$7.12{\textrm{e}}{-3} \pm4.37{\textrm{e}}{-4} $ &$1.26{\textrm{e}}{-3} \pm3.13{\textrm{e}}{-4}$\\
 & MOEIM (\objI-\objS-\objB) & $2.54{\textrm{e}}{-2} \pm1.25{\textrm{e}}{-4} $ &$1.10{\textrm{e}}{-2} \pm7.67{\textrm{e}}{-5} $ &$1.34{\textrm{e}}{-2} \pm2.39{\textrm{e}}{-4} $ & $\cellcolor{green!80}\mathbf{2.10{\textrm{e}}{-2} \pm2.64{\textrm{e}}{-5}}$ &$7.39{\textrm{e}}{-3} \pm7.15{\textrm{e}}{-5} $ & $1.22{\textrm{e}}{-3} \pm4.21{\textrm{e}}{-5}$ &$8.68{\textrm{e}}{-2} \pm6.38{\textrm{e}}{-4} $ &$5.58{\textrm{e}}{-2} \pm4.37{\textrm{e}}{-4} $ &$4.92{\textrm{e}}{-2} \pm2.79{\textrm{e}}{-4} $ &$6.69{\textrm{e}}{-2} \pm3.27{\textrm{e}}{-4} $ &$8.77{\textrm{e}}{-3} \pm5.66{\textrm{e}}{-4} $ &\underline{$1.88{\textrm{e}}{-3} \pm1.59{\textrm{e}}{-4}$} \\
 & MOEIM (\objI-\objS-\objT) & $2.70{\textrm{e}}{-2} \pm1.25{\textrm{e}}{-4} $ &$1.19{\textrm{e}}{-2} \pm1.40{\textrm{e}}{-4} $ &$1.16{\textrm{e}}{-2} \pm3.89{\textrm{e}}{-4} $ &$2.05{\textrm{e}}{-2} \pm7.87{\textrm{e}}{-5} $ &$\cellcolor{green!80}\mathbf{8.29{\textrm{e}}{-3} \pm1.06{\textrm{e}}{-4}}$ & $1.09{\textrm{e}}{-3} \pm6.06{\textrm{e}}{-5}$ &\underline{$9.42{\textrm{e}}{-2} \pm3.14{\textrm{e}}{-4}$} &\underline{$6.03{\textrm{e}}{-2} \pm3.03{\textrm{e}}{-4}$} &$4.83{\textrm{e}}{-2} \pm4.50{\textrm{e}}{-4} $ &$\cellcolor{green!80}\mathbf{6.80{\textrm{e}}{-2} \pm1.40{\textrm{e}}{-4}}$ &\underline{$9.35{\textrm{e}}{-3} \pm3.32{\textrm{e}}{-4}$} & $1.45{\textrm{e}}{-3} \pm1.26{\textrm{e}}{-4}$ \\
 & MOEIM (all) & $2.48{\textrm{e}}{-2} \pm1.12{\textrm{e}}{-4} $ &$1.16{\textrm{e}}{-2} \pm4.41{\textrm{e}}{-5} $ &$1.40{\textrm{e}}{-2} \pm1.03{\textrm{e}}{-4} $ &$2.06{\textrm{e}}{-2} \pm4.15{\textrm{e}}{-5} $ &$7.96{\textrm{e}}{-3} \pm2.09{\textrm{e}}{-4} $ &$\cellcolor{red!70}\mathbf{1.49{\textrm{e}}{-3} \pm4.30{\textrm{e}}{-5}}$ &$8.46{\textrm{e}}{-2} \pm3.15{\textrm{e}}{-4} $ &$5.45{\textrm{e}}{-2} \pm2.83{\textrm{e}}{-4} $ &$5.02{\textrm{e}}{-2} \pm7.96{\textrm{e}}{-4} $ &$6.52{\textrm{e}}{-2} \pm3.54{\textrm{e}}{-4} $ &$\cellcolor{green!80}\mathbf{1.03{\textrm{e}}{-2} \pm5.22{\textrm{e}}{-4}}$ &$\cellcolor{red!70}\mathbf{2.37{\textrm{e}}{-3} \pm1.87{\textrm{e}}{-4}}$ \\
 \noalign{\bigskip}
 \bottomrule
 
 \end{tabular}
 \end{adjustbox}
\end{table*}

\section{Results}
\label{sec:results}

We present now the results of the two experimental settings, followed by a detailed statistical analysis. We conclude this section with an analysis of the correlation between the objective functions.

\subsection{Setting \setone: \algname vs.\ optimization methods}
\label{sec:setting1}
The first experimental setting aims to analyze the performance of \algname w.r.t. the the existing optimization methods for IM. These experiments are designed to understand if in the classical (\objI-\objS) setting \algname can produce better results, but also to analyze the performance of \algname when considering additional objective functions. We tested \algname in six different optimization settings, namely the (\objI-\objS) setting used in \cite{bucur2018improving}, four different settings where each of the additional objective functions (i.e., \objC, \objF, \objB, \objT) are added to (\objI-\objS) in the optimization process and, to conclude, a setting called ``(all)'' where all the six objective functions are optimized together.
The baselines are computed as proposed in the original papers, i.e., GDD \cite{wang2016maximizing} and CELF \cite{leskovec2007cost} are executed under a seed set size constraint, while MOEA is executed on the base case (\objI-\objS), as in \cite{bucur2018improving}.
The results of such experiments are available in \Cref{tab:results}, where for each algorithm we extracted the hypervolume \cite{shang2020survey} w.r.t. all the six optimization settings mentioned above. The hypervolume is computed over the normalized values of each objective function as: $\text{\objI} \in [0,|\mathcal{V}|]$, 
$\text{\objS} \in [0,k]$,
$\text{\objC} \in [0,1]$,
$\text{\objF} \in [0,1]$,
$\text{\objB} \in [0,b]$, where $b=\sum_i^k d_{v_i}^{out}$ with $v$ being the list of the top-$k$ nodes by out-degree, and $\text{\objT} \in [0,\tau]$.


Overall, \algname (in any of its $6$ configurations) turns out to outperform the three baselines in $61$ cases out of $72$ (namely, 6 datasets $\times$ 2 propagation models $\times$ 6 combinations of objectives), see the colored cells. In $10$ of out of the remaining $11$ cases, one of the $6$ \algname configurations turns out to be the second best (the only case where this does not happen is when calculating the hypervolume in the (\objI-\objS-\objB) space on the email-eu-Core dataset, where GDD and CELF are respectively the first and second best algorithms).
More interesting, we notice that our proposed \algname turns out not to be the best only in some cases that present a pattern: 1) on sparse graphs (such as gnutella, CA-HepTh, and lastfm, see \Cref{tab:dataset}); and 2) in the (\objI,\objS) and the \objB settings with the WC model. In the first case, our proposed algorithm struggles since its design is based on smart initialization and graph-aware operators, which require strong connectivity of the graphs and high variance between node's out-degrees to be effective. Hence, in the sparser graphs, even if the solutions are at least always the second best, the low connectivity and low variance of out-degree leads to suboptimal results. The case of \objB is instead most likely due to smart initialization, which by design selects in the seed set the nodes with the highest out-degree, while the other baselines do not have this kind of bias on the seed set initialization.

The final consideration concerns the combination of objective functions. Ideally, when optimizing a third objective function (i.e., (\objI-\objS) plus either \objC, \objF, \objB or \objS, see the ``Algorithm'' column in \Cref{tab:results}), the hypervolume w.r.t. (\objI-\objS) plus the third objective (top row of the table) should provide the highest results. This happens in several cases, as in the (\objI-\objS), ``all'', and when optimizing \objC, \objF, and \objB. On the other hand, when optimizing \objT, the best hypervolume turns out to be related to (\objI-\objS-\objB) for the ``all'' setting, shading lights to possible correlation between objective functions, that we discuss below. The analysis of the correlation is also motivated to explain why, when the additional objective functions are added to (\objI-\objS), the hypervolume of (\objI-\objS) decreases.

\subsection{Setting \settwo: \algname vs.\ DeepIM}
\label{sec:setting2}
\begin{figure*}[!t]
 \centering
 \includegraphics[width=1\textwidth]{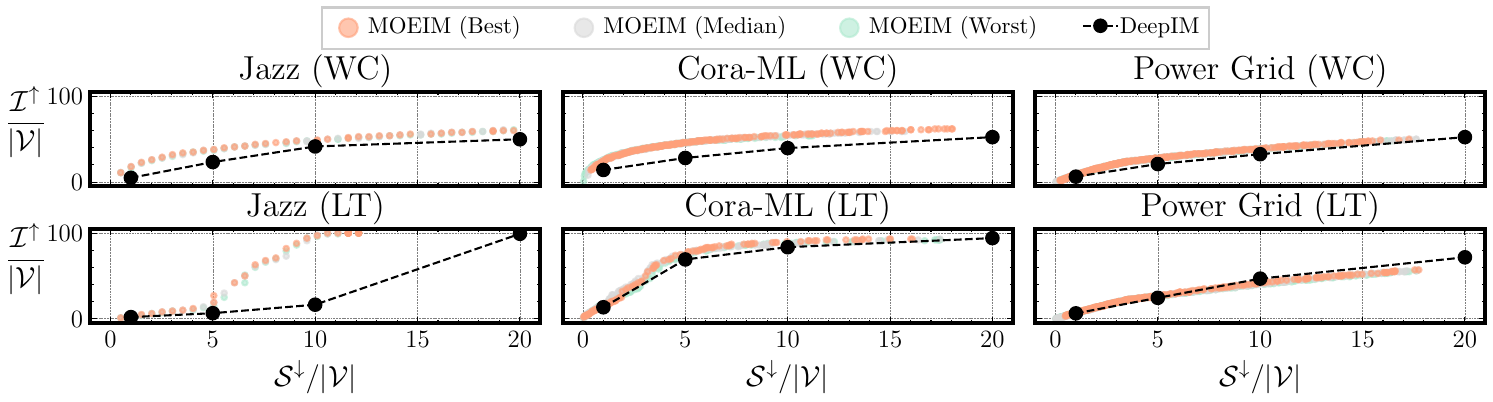}
 \caption{Non-dominated solutions found \algname and DeepIM \cite{ling2023deep}. For \algname, we provide the worst, median and best sets of non-dominated solutions found across $10$ runs. To allow for a direct comparison with the results reported in \cite{ling2023deep}, the x-axis and y-axis show, respectively, the seed set size \objS and the final influence \objI, both normalized w.r.t. the size of each network, $|\mathcal{V}|$. For DeepIM, we show the results available in the original paper, where they are reported separately for each value of \objS/$|\mathcal{V}|$.}
 \label{fig:ea_ml}
\end{figure*}

In this section, we compare \algname with DeepIM \cite{ling2023deep}, a state-of-the-art DL approach for the IM problem. DeepIM is a Graph Neural Network (GNN)-based Diffusion Model designed to translate the discrete optimization IM problem into a continuous space, and then solve it by means of DL. Since in \cite{ling2023deep} it has been empirically proven that DeepIM outperforms other DL-based algorithms specifically designed for IM (such as PIANO \cite{li2022piano} and ToupleGDD \cite{chen2023touplegdd}), we directly take the numerical results reported in that paper, considering the three different datasets mentioned above (Jazz, Cora-ML, and Power Grid) under two different diffusion models (IC and LT). 
In this setting, we run our \algname only on the (\objI-\objS) setting, for which the DeepIM results can be derived directly from the original paper. Furthermore, for \algname we use the same hyperparameters as in the first experimental setting. 
For the sake of visualization, in \Cref{fig:ea_ml} we show the non-dominated solutions found by \algname (in the best, median and worst case across the $10$ available runs) along with the solutions found by DeepIM reported in \cite{ling2023deep}\footnote{As DeepIM is a single-objective algorithm, the four points shown in each subfigure in \Cref{fig:ea_ml} are related to four independent executions of DeepIM for different values of the seed set size, namely $1\%$, $5\%$, $10\%$, and $20\%$ of the graph size. The values used for the figure have been extracted from \cite{ling2023deep}.}.

As can be seen from the figure, \algname is able to outperform DeepIM on almost all propagation models and datasets. The only exception is the Power Grid dataset with LT, where the results are almost identical until a seed set size of $k=10\%$ of the graph size, after which DeepIM shows superior performance. As discussed in the previous experimental setting, this trend is most likely due to the fact that this dataset is the sparsest one (i.e., the one with the lowest avg. out-degree) among those tested in this setting, which makes the initialization and graph-aware operators less effective.

On the other hand the superiority of \algname is particularly clear on the Jazz dataset with the LT model, where it only needs a seed set size of $10\%$ of the graph size to influence the whole network, while DeepIM achieves the same results using twice the number of nodes.
Overall, it appears that in most cases \algname tends to provide better performance than DeepIM, which as said is currently the best DL-based approach available in the literature.

\begin{table}[t!]
 \centering
 \caption{Results of the statistical tests, performed by using the Holm-Bonferroni procedure \cite{holm1979simple}, using ``\algname (all)'' as reference (Rank=8.82). Note that, for simplicity of notation, we omit ``\algname'' from the other settings of our algorithm. ``Rank'' refers to the ranking of the algorithm (the higher, the better). ``$z$''
 is the statistic computed on the rank, ``$p$'' is the $p$-value, computed as the value of the cumulative normal distribution related to $z$. $\alpha/j$ is the significance threshold adjusted for each row of the table. ``Reject'' indicates if the sequential null hypothesis (of equivalence between the reference and the algorithm on the row) can be rejected or not.}
 \label{tab:holm-bonferroni_test}
 \resizebox{\columnwidth}{!}{
 \begin{tabular}{lccccc}
 \toprule
 \textbf{Method} & \textbf{Rank} & $z$ & $p$ & $\alpha/j$ & \textbf{Rejected} \\
 \midrule
 \objI-\objS-\objB & 6.56e+00 & -4.71e+00 & 1.21e-06 & 5.00e-02 & True \\
 \objI-\objS-\objT & 6.52e+00 & -4.83e+00 & 6.76e-07 & 2.50e-02 & True \\
 \objI-\objS-\objF & 4.97e+00 & -9.22e+00 & 1.54e-20 & 1.67e-02 & True \\
 \objI-\objS-\objC & 3.55e+00 & -1.32e+01 & 3.24e-40 & 1.25e-02 & True \\
 GGD & 3.46e+00 & -1.35e+01 & 9.96e-42 & 1.00e-02 & True \\
 CELF & 3.36e+00 & -1.38e+01 & 2.31e-43 & 8.33e-03 & True \\
 MOEA & 2.27e+00 & -1.69e+01 & 5.01e-64 & 7.14e-03 & True \\
 \objI-\objS & 2.15e+00 & -1.72e+01 & 1.79e-66 & 6.25e-03 & True \\
 \bottomrule
 \end{tabular}
 }
\end{table}

\subsection{Statistical analysis}
To ensure the statistical relevance of our results, we performed a statistical comparison on the hypervolumes achieved by the algorithms under comparison in the first experimental setting \setone, by using the Holm-Bonferroni procedure \cite{holm1979simple}, with significance threshold $\alpha=0.05$ and the null hypothesis of statistical equivalence. 
From this test (whose results are shown in \Cref{tab:holm-bonferroni_test}), it emerges that the best-performing method is \algname in the ``all'' setting (i.e., the one considering all six objective functions), which allows rejecting the null hypothesis in all the cases.
This is clearly a confirmation of an expected result, as we expect the hypervolume to be maximized when we optimize all the objectives simultaneously.

\begin{table}[ht!]
 \centering
 \caption{Results of the statistical tests using the Tukey HSD procedure \cite{tukey1949comparing} for pairwise comparisons. Note that, for simplicity of notation, we omit ``\algname'' from the settings of our algorithm. ``Mean difference'' refers to the mean of the difference on the hypervolume achieved by the methods in each pairwise comparison, ``$p\textrm{-adj}$'' refers to the adjusted $p$-value, ``Lower CI'' and ``Upper CI'' refer to the lower and upper bound for the $95\%$ CI (Confidence Interval). ``Reject'' indicates if the null hypothesis (of statistical equivalence between Methods 1 and 2) can be rejected or not. We highlight the cases for which the hypothesis can be rejected.}
 \label{tab:tukey_test}
 \resizebox{\columnwidth}{!}{
\begin{tabular}{llccccc}
 \toprule
 \textbf{Method 1} & \textbf{Method 2} & \textbf{Mean difference} & $p\textrm{-adj}$ & \textbf{Lower CI} & \textbf{Upper CI} & \textbf{Rejected} \\
 \midrule
 \rowcolor{rejectColor} \objI-\objS & \objI-\objS-\objB & 6.90e-3 & 3.00e-4 & 2.10e-3 & 1.16e-2 & True \\
 \objI-\objS & \objI-\objS-\objC & 1.40e-3 & 9.93e-1 & -3.40e-3 & 6.20e-3 & False \\
 \objI-\objS & \objI-\objS-\objF & 2.10e-3 & 9.06e-1 & -2.70e-3 & 6.90e-3 & False \\
 \rowcolor{rejectColor} \objI-\objS & \objI-\objS-\objT & 7.70e-3 & 0.00e+0 & 3.00e-3 & 1.25e-2 & True \\
 \rowcolor{rejectColor} \objI-\objS & all & 1.20e-2 & 0.00e+0 & 7.30e-3 & 1.68e-2 & True \\
 \objI-\objS & CELF & 2.40e-3 & 9.80e-1 & -4.60e-3 & 9.40e-3 & False \\
 \objI-\objS & GGD & 3.80e-3 & 2.62e-1 & -1.00e-3 & 8.50e-3 & False \\
 \objI-\objS & MOEA & 0.00e+0 & 1.00e+0 & -4.70e-3 & 4.80e-3 & False \\
 \rowcolor{rejectColor} \objI-\objS-\objB & \objI-\objS-\objC & -5.50e-3 & 1.09e-2 & -1.03e-2 & -7.00e-4 & True \\
 \objI-\objS-\objB & \objI-\objS-\objF & -4.80e-3 & 5.11e-2 & -9.50e-3 & 0.00e+0 & False \\
 \objI-\objS-\objB & \objI-\objS-\objT & 9.00e-4 & 1.00e+0 & -3.90e-3 & 5.60e-3 & False \\
 \rowcolor{rejectColor} \objI-\objS-\objB & all & 5.20e-3 & 2.28e-2 & 4.00e-4 & 9.90e-3 & True \\
 \objI-\objS-\objB & CELF & -4.50e-3 & 5.53e-1 & -1.15e-2 & 2.50e-3 & False \\
 \objI-\objS-\objB & GGD & -3.10e-3 & 5.19e-1 & -7.90e-3 & 1.60e-3 & False \\
 \rowcolor{rejectColor} \objI-\objS-\objB & MOEA & -6.90e-3 & 3.00e-4 & -1.16e-2 & -2.10e-3 & True \\
 \objI-\objS-\objC & \objI-\objS-\objF & 7.00e-4 & 1.00e+0 & -4.00e-3 & 5.50e-3 & False \\
 \rowcolor{rejectColor} \objI-\objS-\objC & \objI-\objS-\objT & 6.40e-3 & 1.30e-3 & 1.60e-3 & 1.11e-2 & True \\
 \rowcolor{rejectColor} \objI-\objS-\objC & all & 1.07e-2 & 0.00e+0 & 5.90e-3 & 1.54e-2 & True \\
 \objI-\objS-\objC & CELF & 1.00e-3 & 1.00e+0 & -6.00e-3 & 8.00e-3 & False \\
 \objI-\objS-\objC & GGD & 2.40e-3 & 8.36e-1 & -2.40e-3 & 7.10e-3 & False \\
 \objI-\objS-\objC & MOEA & -1.40e-3 & 9.94e-1 & -6.10e-3 & 3.40e-3 & False \\
 \rowcolor{rejectColor} \objI-\objS-\objF & \objI-\objS-\objT & 5.60e-3 & 8.10e-3 & 8.00e-4 & 1.04e-2 & True \\
 \rowcolor{rejectColor} \objI-\objS-\objF & all & 9.90e-3 & 0.00e+0 & 5.10e-3 & 1.47e-2 & True \\
 \objI-\objS-\objF & CELF & 3.00e-4 & 1.00e+0 & -6.80e-3 & 7.30e-3 & False \\
 \objI-\objS-\objF & GGD & 1.60e-3 & 9.79e-1 & -3.10e-3 & 6.40e-3 & False \\
 \objI-\objS-\objF & MOEA & -2.10e-3 & 9.12e-1 & -6.90e-3 & 2.70e-3 & False \\
 \objI-\objS-\objT & all & 4.30e-3 & 1.17e-1 & -5.00e-4 & 9.10e-3 & False \\
 \objI-\objS-\objT & CELF & -5.40e-3 & 3.03e-1 & -1.24e-2 & 1.70e-3 & False \\
 \objI-\objS-\objT & GGD & -4.00e-3 & 1.89e-1 & -8.80e-3 & 8.00e-4 & False \\
 \rowcolor{rejectColor} \objI-\objS-\objT & MOEA & -7.70e-3 & 0.00e+0 & -1.25e-2 & -2.90e-3 & True \\
 \rowcolor{rejectColor} all & CELF & -9.70e-3 & 7.00e-4 & -1.67e-2 & -2.60e-3 & True \\
 \rowcolor{rejectColor} all & GGD & -8.30e-3 & 0.00e+0 & -1.31e-2 & -3.50e-3 & True \\
 \rowcolor{rejectColor} all & MOEA & -1.20e-2 & 0.00e+0 & -1.68e-2 & -7.20e-3 & True \\
 CELF & GGD & 1.40e-3 & 1.00e+0 & -5.70e-3 & 8.40e-3 & False \\
 CELF & MOEA & -2.40e-3 & 9.81e-1 & -9.40e-3 & 4.70e-3 & False \\
 GGD & MOEA & -3.70e-3 & 2.71e-1 & -8.50e-3 & 1.00e-3 & False \\
\bottomrule
\end{tabular}
}
\end{table}

Moreover, to statistically assess the pairwise comparisons among the algorithms, we also performed a Tukey HSD test \cite{tukey1949comparing} for multiple comparisons, keeping the same significance threshold and null hypothesis.
These results (presented in \Cref{tab:tukey_test}) show that, while in \Cref{tab:results} all the \algname settings seem to outperform the baseline methods, only the ``all'' setting reliably does so.
It is worth noting that this may result be due to the fact that, since this test is based on the actual Hypervolumes (instead of their ranking, as in the Holm-Bonferroni test) and we are aggregating the results obtained across several datasets, the standard deviation of the Hypervolumes obtained with the other \algname settings proposed in this paper is too high to lead to statistically significant results.

Finally, it is important to note that we could not perform any statistical test between our methods and DeepIM considered in the second experimental setting \settwo, since the DeepIM results have been taken directly from the original paper without raw data availability.
\vspace{-.5cm}


\begin{figure*}[!t]
 \centering
 \includegraphics[width=1\textwidth]{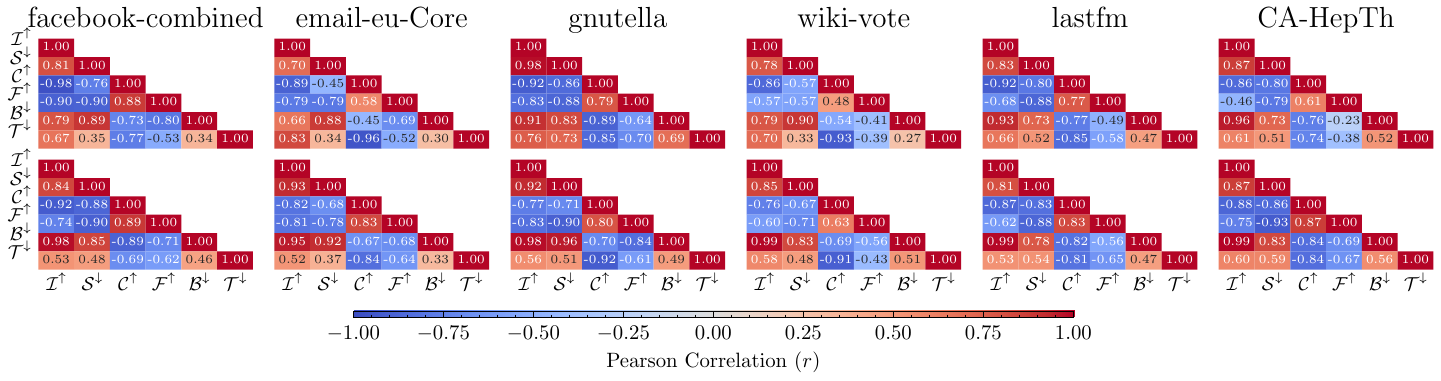}
 \caption{Pearson correlation among the objectives (\objI: Influence; \objS: Seed set size; \objC: Communities; \objF: Fairness; \objB: Budget; \objT: Time). This correlation has been computed over the results of the 10 runs available for the ``\algname (all)'' setting (see \Cref{tab:results} for details). Top row: IC model, bottom row: WC model.}
 \label{fig:correlation_IC}
\end{figure*}

\subsection{Analysis of objective function correlation}
We conclude our analysis studying the correlation between the objective functions. This study has been conducted to better understand the hypervolume trend on \objT of \Cref{tab:results} and explain the reason why the hypervolume computed over (\objI-\objS) decreases when adding objective functions.
\Cref{fig:correlation_IC} displays the Pearson correlation among the proposed objective functions. Such correlation has been computed separately for each $\langle$dataset, propagation model$\rangle$ combination on the results of \algname (all). 
For this analysis, each entry corresponds to the vector of the normalized objectives (see \Cref{sec:setting1}) for each non-dominated solution found across the $10$ runs of \algname (all).
The analysis clearly shows how all the objective functions added to the base setting objectives (i.e., \objI and \objS) are negatively correlated with them. In fact, \objC and \objF are negatively correlated (w.r.t. to both \objI and \objS), while \objB and \objT are positively correlated (always w.r.t. \objI and \objS), but these latter two objectives are minimized. The highest correlation turns out to be between \objB and \objI, since as expected selecting (in the seed set) nodes with high out-degree leads to improving the overall influence spread, and between \objS and \objI. Of note, the correlation between \objC and \objF turns out to be extremely high, since such objectives, even tough formulated differently, share the same spirit of fair allocation on either the final influence or the seed set allocation. 
To conclude, the reason why, in some cases, the hypervolume in the \objB or ``all'' settings is higher w.r.t. (\objI-\objS-\objT) is due to the positive correlation between \objT with both \objB and \objS.


\section{Conclusions}
\label{sec:conclusions}
We presented a new setting for the IM problem, where up to six objective functions are considered in the optimization process. For solving such a task, we proposed \algname, a MOEA designed to leverage graph information thanks to the use of graph-aware operators and a smart initialization of the initial population. The numerical results confirm the effectiveness of the proposed approach w.r.t. the state of the art, including a recently proposed DL-based method. Moreover, our results allowed us to investigate the correlation objectives, providing insightful information on IM.

The approach presented in this work opens up several possibilities to expand the IM problem even further, as new objectives can be easily added to our codebase. Moreover, while in this work we mainly focused on static graphs and propagation models taken independently, another possibility for future works could be to conduct similar multi-objective studies on dynamic graphs, or to aim at assessing the solutions robustness w.r.t. multiple propagation models.




\clearpage
\balance

\bibliographystyle{ACM-Reference-Format}
\bibliography{bib-file}


\begin{thebibliography}{58}


\ifx \showCODEN    \undefined \def \showCODEN     #1{\unskip}     \fi
\ifx \showDOI      \undefined \def \showDOI       #1{#1}\fi
\ifx \showISBNx    \undefined \def \showISBNx     #1{\unskip}     \fi
\ifx \showISBNxiii \undefined \def \showISBNxiii  #1{\unskip}     \fi
\ifx \showISSN     \undefined \def \showISSN      #1{\unskip}     \fi
\ifx \showLCCN     \undefined \def \showLCCN      #1{\unskip}     \fi
\ifx \shownote     \undefined \def \shownote      #1{#1}          \fi
\ifx \showarticletitle \undefined \def \showarticletitle #1{#1}   \fi
\ifx \showURL      \undefined \def \showURL       {\relax}        \fi
\providecommand\bibfield[2]{#2}
\providecommand\bibinfo[2]{#2}
\providecommand\natexlab[1]{#1}
\providecommand\showeprint[2][]{arXiv:#2}

\bibitem[Ali et~al\mbox{.}(2023)]%
        {ali2021fairness}
\bibfield{author}{\bibinfo{person}{Junaid Ali}, \bibinfo{person}{Mahmoudreza
  Babaei}, \bibinfo{person}{Abhijnan Chakraborty}, \bibinfo{person}{Baharan
  Mirzasoleiman}, \bibinfo{person}{Krishna Gummadi}, {and}
  \bibinfo{person}{Adish Singla}.} \bibinfo{year}{{2023}}\natexlab{}.
\newblock \showarticletitle{{On the fairness of time-critical influence
  maximization in social networks}}.
\newblock \bibinfo{journal}{\emph{{IEEE Transactions on Knowledge and Data
  Engineering}}} \bibinfo{volume}{{35}}, \bibinfo{number}{{3}}
  (\bibinfo{year}{{2023}}), \bibinfo{pages}{2875--2886}.
\newblock


\bibitem[Becker et~al\mbox{.}(2022)]%
        {becker2022fairness}
\bibfield{author}{\bibinfo{person}{Ruben Becker}, \bibinfo{person}{Gianlorenzo
  D’angelo}, \bibinfo{person}{Sajjad Ghobadi}, {and} \bibinfo{person}{Hugo
  Gilbert}.} \bibinfo{year}{{2022}}\natexlab{}.
\newblock \showarticletitle{{Fairness in influence maximization through
  randomization}}.
\newblock \bibinfo{journal}{\emph{{Journal of Artificial Intelligence
  Research}}}  \bibinfo{volume}{{73}} (\bibinfo{year}{{2022}}),
  \bibinfo{pages}{{1251--1283}}.
\newblock


\bibitem[{Benedek Rozemberczki and Rik Sarkar}(2020)]%
        {feather}
\bibfield{author}{\bibinfo{person}{{Benedek Rozemberczki and Rik Sarkar}}.}
  \bibinfo{year}{{2020}}\natexlab{}.
\newblock \showarticletitle{{Characteristic Functions on Graphs: Birds of a
  Feather, from Statistical Descriptors to Parametric Models}}. In
  \bibinfo{booktitle}{\emph{{International Conference on Information and
  Knowledge Management}}}. \bibinfo{publisher}{{ACM}}, \bibinfo{address}{{New
  York, NY, USA}}, \bibinfo{pages}{{1325–1334}}.
\newblock


\bibitem[{Bian, Song and Guo, Qintian and Wang, Sibo and Yu, Jeffrey
  Xu}(2020)]%
        {bian2020efficient}
\bibfield{author}{\bibinfo{person}{{Bian, Song and Guo, Qintian and Wang, Sibo
  and Yu, Jeffrey Xu}}.} \bibinfo{year}{{2020}}\natexlab{}.
\newblock \showarticletitle{{Efficient algorithms for budgeted influence
  maximization on massive social networks}}.
\newblock \bibinfo{journal}{\emph{{Proceedings of the VLDB Endowment}}}
  \bibinfo{volume}{{13}}, \bibinfo{number}{{9}} (\bibinfo{year}{{2020}}),
  \bibinfo{pages}{{1498--1510}}.
\newblock


\bibitem[Biswas et~al\mbox{.}(2022a)]%
        {biswas2022improved}
\bibfield{author}{\bibinfo{person}{Tarun~K Biswas}, \bibinfo{person}{Alireza
  Abbasi}, {and} \bibinfo{person}{Ripon~K Chakrabortty}.}
  \bibinfo{year}{{2022}}\natexlab{a}.
\newblock \showarticletitle{{An improved clustering based multi-objective
  evolutionary algorithm for influence maximization under variable-length
  solutions}}.
\newblock \bibinfo{journal}{\emph{{Knowledge-Based Systems}}}
  \bibinfo{volume}{{256}} (\bibinfo{year}{{2022}}), \bibinfo{pages}{{109856}}.
\newblock


\bibitem[Biswas et~al\mbox{.}(2022b)]%
        {biswas2022multi}
\bibfield{author}{\bibinfo{person}{Tarun~K Biswas}, \bibinfo{person}{Alireza
  Abbasi}, {and} \bibinfo{person}{Ripon~K Chakrabortty}.}
  \bibinfo{year}{{2022}}\natexlab{b}.
\newblock \showarticletitle{{Multi-Objective Influence Maximization Under
  Varying-Size Solutions and Constraints}}. In
  \bibinfo{booktitle}{\emph{{IEEE/ACM International Conference on Advances in
  Social Networks Analysis and Mining}}}. \bibinfo{publisher}{{IEEE}},
  \bibinfo{address}{{New York, NY, USA}}, \bibinfo{pages}{{285--292}}.
\newblock


\bibitem[Bucur and Iacca(2016)]%
        {bucur2016influence}
\bibfield{author}{\bibinfo{person}{Doina Bucur} {and} \bibinfo{person}{Giovanni
  Iacca}.} \bibinfo{year}{{2016}}\natexlab{}.
\newblock \showarticletitle{{Influence maximization in social networks with
  genetic algorithms}}. In \bibinfo{booktitle}{\emph{{Applications of
  Evolutionary Computation}}}. \bibinfo{publisher}{{Springer}},
  \bibinfo{address}{{Cham, Switzerland}}, \bibinfo{pages}{{379--392}}.
\newblock


\bibitem[Bucur et~al\mbox{.}(2017)]%
        {bucur2017multi}
\bibfield{author}{\bibinfo{person}{Doina Bucur}, \bibinfo{person}{Giovanni
  Iacca}, \bibinfo{person}{Andrea Marcelli}, \bibinfo{person}{Giovanni
  Squillero}, {and} \bibinfo{person}{Alberto Tonda}.}
  \bibinfo{year}{{2017}}\natexlab{}.
\newblock \showarticletitle{{Multi-objective evolutionary algorithms for
  influence maximization in social networks}}. In
  \bibinfo{booktitle}{\emph{{Applications of Evolutionary Computation}}}.
  \bibinfo{publisher}{{Springer}}, \bibinfo{address}{{Cham, Switzerland}},
  \bibinfo{pages}{{221--233}}.
\newblock


\bibitem[Bucur et~al\mbox{.}(2018)]%
        {bucur2018improving}
\bibfield{author}{\bibinfo{person}{Doina Bucur}, \bibinfo{person}{Giovanni
  Iacca}, \bibinfo{person}{Andrea Marcelli}, \bibinfo{person}{Giovanni
  Squillero}, {and} \bibinfo{person}{Alberto Tonda}.}
  \bibinfo{year}{{2018}}\natexlab{}.
\newblock \showarticletitle{{Improving multi-objective evolutionary influence
  maximization in social networks}}. In \bibinfo{booktitle}{\emph{{Applications
  of Evolutionary Computation}}}. \bibinfo{publisher}{{Springer}},
  \bibinfo{address}{{Cham, Switzerland}}, \bibinfo{pages}{{117--124}}.
\newblock


\bibitem[{Chawla, Aviral and Cheney, Nick}(2023)]%
        {chawla2023neighbor}
\bibfield{author}{\bibinfo{person}{{Chawla, Aviral and Cheney, Nick}}.}
  \bibinfo{year}{{2023}}\natexlab{}.
\newblock \showarticletitle{{Neighbor-Hop Mutation for Genetic Algorithm in
  Influence Maximization}}. In \bibinfo{booktitle}{\emph{{Conference on Genetic
  and Evolutionary Computation Companion}}}. \bibinfo{publisher}{{ACM}},
  \bibinfo{address}{{New York, NY, USA}}, \bibinfo{pages}{{187--190}}.
\newblock


\bibitem[Chen et~al\mbox{.}(2023)]%
        {chen2023touplegdd}
\bibfield{author}{\bibinfo{person}{Tiantian Chen}, \bibinfo{person}{Siwen Yan},
  \bibinfo{person}{Jianxiong Guo}, {and} \bibinfo{person}{Weili Wu}.}
  \bibinfo{year}{{2023}}\natexlab{}.
\newblock \showarticletitle{{ToupleGDD: A Fine-Designed Solution of Influence
  Maximization by Deep Reinforcement Learning}}.
\newblock \bibinfo{journal}{\emph{{IEEE Transactions on Computational Social
  Systems}}}  \bibinfo{volume}{{(early access)}} (\bibinfo{year}{{2023}}),
  \bibinfo{pages}{{1--12}}.
\newblock


\bibitem[Chen et~al\mbox{.}(2012)]%
        {chen2012time}
\bibfield{author}{\bibinfo{person}{Wei Chen}, \bibinfo{person}{Wei Lu}, {and}
  \bibinfo{person}{Ning Zhang}.} \bibinfo{year}{{2012}}\natexlab{}.
\newblock \showarticletitle{{Time-critical influence maximization in social
  networks with time-delayed diffusion process}}. In
  \bibinfo{booktitle}{\emph{{AAAI Conference on Artificial Intelligence}}},
  Vol.~\bibinfo{volume}{{26}}. \bibinfo{publisher}{{Association for the
  Advancement of Artificial Intelligence}}, \bibinfo{address}{{Washington DC,
  USA}}, \bibinfo{pages}{{591--598}}.
\newblock


\bibitem[{Chen, Wei and Wang, Yajun and Yang, Siyu}(2009)]%
        {chen2009efficient}
\bibfield{author}{\bibinfo{person}{{Chen, Wei and Wang, Yajun and Yang,
  Siyu}}.} \bibinfo{year}{{2009}}\natexlab{}.
\newblock \showarticletitle{{Efficient influence maximization in social
  networks}}. In \bibinfo{booktitle}{\emph{{ACM SIGKDD International Conference
  on Knowledge Discovery and Data Mining}}}. \bibinfo{publisher}{{ACM}},
  \bibinfo{address}{{New York, NY, USA}}, \bibinfo{pages}{{199--208}}.
\newblock


\bibitem[Cunegatti et~al\mbox{.}(2022)]%
        {cunegatti2022large}
\bibfield{author}{\bibinfo{person}{Elia Cunegatti}, \bibinfo{person}{Giovanni
  Iacca}, {and} \bibinfo{person}{Doina Bucur}.}
  \bibinfo{year}{{2022}}\natexlab{}.
\newblock \showarticletitle{{Large-scale multi-objective influence maximisation
  with network downscaling}}. In \bibinfo{booktitle}{\emph{{Parallel Problem
  Solving from Nature}}}. \bibinfo{publisher}{{Springer}},
  \bibinfo{address}{{Cham, Switzerland}}, \bibinfo{pages}{{207--220}}.
\newblock


\bibitem[{Deb, K. and Pratap, A. and Agarwal, S. and Meyarivan, T.}(2002)]%
        {996017}
\bibfield{author}{\bibinfo{person}{{Deb, K. and Pratap, A. and Agarwal, S. and
  Meyarivan, T.}}} \bibinfo{year}{{2002}}\natexlab{}.
\newblock \showarticletitle{{A fast and elitist multiobjective genetic
  algorithm: NSGA-II}}.
\newblock \bibinfo{journal}{\emph{{IEEE Transactions on Evolutionary
  Computation}}} \bibinfo{volume}{{6}}, \bibinfo{number}{{2}}
  (\bibinfo{year}{{2002}}), \bibinfo{pages}{{182--197}}.
\newblock


\bibitem[Farnad et~al\mbox{.}(2020)]%
        {farnad2020unifying}
\bibfield{author}{\bibinfo{person}{Golnoosh Farnad}, \bibinfo{person}{Behrouz
  Babaki}, {and} \bibinfo{person}{Michel Gendreau}.}
  \bibinfo{year}{{2020}}\natexlab{}.
\newblock \showarticletitle{{A unifying framework for fairness-aware influence
  maximization}}. In \bibinfo{booktitle}{\emph{{The Web Conference
  Companion}}}. \bibinfo{publisher}{{ACM}}, \bibinfo{address}{{New York, NY,
  USA}}, \bibinfo{pages}{{714--722}}.
\newblock


\bibitem[Feng et~al\mbox{.}(2023)]%
        {feng2023influence}
\bibfield{author}{\bibinfo{person}{Yuting Feng}, \bibinfo{person}{Ankitkumar
  Patel}, \bibinfo{person}{Bogdan Cautis}, {and} \bibinfo{person}{Hossein
  Vahabi}.} \bibinfo{year}{{2023}}\natexlab{}.
\newblock \bibinfo{title}{{Influence Maximization with Fairness at Scale}}.
\newblock
\newblock
\newblock
\shownote{{arXiv preprint arXiv:2306.01587}}.


\bibitem[{Giovanni Iacca and Kateryna Konotopska and Doina Bucur and Alberto
  Tonda}(2021)]%
        {IACCA2021100107}
\bibfield{author}{\bibinfo{person}{{Giovanni Iacca and Kateryna Konotopska and
  Doina Bucur and Alberto Tonda}}.} \bibinfo{year}{{2021}}\natexlab{}.
\newblock \showarticletitle{{An evolutionary framework for maximizing influence
  propagation in social networks}}.
\newblock \bibinfo{journal}{\emph{{Software Impacts}}}  \bibinfo{volume}{{9}}
  (\bibinfo{year}{{2021}}), \bibinfo{pages}{{100107}}.
\newblock


\bibitem[{Gleiser, Pablo M and Danon, Leon}(2003)]%
        {gleiser2003community}
\bibfield{author}{\bibinfo{person}{{Gleiser, Pablo M and Danon, Leon}}.}
  \bibinfo{year}{{2003}}\natexlab{}.
\newblock \showarticletitle{{Community structure in jazz}}.
\newblock \bibinfo{journal}{\emph{{Advances in complex systems}}}
  \bibinfo{volume}{{6}}, \bibinfo{number}{{04}} (\bibinfo{year}{{2003}}),
  \bibinfo{pages}{{565--573}}.
\newblock


\bibitem[Gong and Guo(2023)]%
        {gong2023influence}
\bibfield{author}{\bibinfo{person}{Hao Gong} {and} \bibinfo{person}{Chunxiang
  Guo}.} \bibinfo{year}{{2023}}\natexlab{}.
\newblock \showarticletitle{{Influence maximization considering fairness: A
  multi-objective optimization approach with prior knowledge}}.
\newblock \bibinfo{journal}{\emph{{Expert Systems with Applications}}}
  \bibinfo{volume}{{214}} (\bibinfo{year}{{2023}}), \bibinfo{pages}{{119138}}.
\newblock


\bibitem[{Gong, Maoguo and Yan, Jianan and Shen, Bo and Ma, Lijia and Cai,
  Qing}(2016)]%
        {gong2016influence}
\bibfield{author}{\bibinfo{person}{{Gong, Maoguo and Yan, Jianan and Shen, Bo
  and Ma, Lijia and Cai, Qing}}.} \bibinfo{year}{{2016}}\natexlab{}.
\newblock \showarticletitle{{Influence maximization in social networks based on
  discrete particle swarm optimization}}.
\newblock \bibinfo{journal}{\emph{{Information Sciences}}}
  \bibinfo{volume}{{367}} (\bibinfo{year}{{2016}}),
  \bibinfo{pages}{{600--614}}.
\newblock


\bibitem[Goyal et~al\mbox{.}(2011)]%
        {goyal2011celf++}
\bibfield{author}{\bibinfo{person}{Amit Goyal}, \bibinfo{person}{Wei Lu}, {and}
  \bibinfo{person}{Laks~VS Lakshmanan}.} \bibinfo{year}{{2011}}\natexlab{}.
\newblock \showarticletitle{{CELF++ optimizing the greedy algorithm for
  influence maximization in social networks}}. In
  \bibinfo{booktitle}{\emph{{International Conference on World Wide Web
  Companion}}}. \bibinfo{publisher}{{ACM}}, \bibinfo{address}{{New York, NY,
  USA}}, \bibinfo{pages}{{47--48}}.
\newblock


\bibitem[{Holm, Sture}(1979)]%
        {holm1979simple}
\bibfield{author}{\bibinfo{person}{{Holm, Sture}}.}
  \bibinfo{year}{{1979}}\natexlab{}.
\newblock \showarticletitle{{A simple sequentially rejective multiple test
  procedure}}.
\newblock \bibinfo{journal}{\emph{{Scandinavian journal of statistics}}}
  \bibinfo{volume}{{6}}, \bibinfo{number}{{2}} (\bibinfo{year}{{1979}}),
  \bibinfo{pages}{{65--70}}.
\newblock


\bibitem[{Jiang, Qingye and Song, Guojie and Gao, Cong and Wang, Yu and Si,
  Wenjun and Xie, Kunqing}(2011)]%
        {jiang2011simulated}
\bibfield{author}{\bibinfo{person}{{Jiang, Qingye and Song, Guojie and Gao,
  Cong and Wang, Yu and Si, Wenjun and Xie, Kunqing}}.}
  \bibinfo{year}{{2011}}\natexlab{}.
\newblock \showarticletitle{{Simulated annealing based influence maximization
  in social networks}}. In \bibinfo{booktitle}{\emph{{AAAI Conference on
  Artificial intelligence}}}, Vol.~\bibinfo{volume}{{25}}.
  \bibinfo{publisher}{{Association for the Advancement of Artificial
  Intelligence}}, \bibinfo{address}{{Washington DC, USA}},
  \bibinfo{pages}{{127--132}}.
\newblock


\bibitem[{Jure Leskovec and Andrej Krevl}(2014)]%
        {snapnets}
\bibfield{author}{\bibinfo{person}{{Jure Leskovec and Andrej Krevl}}.}
  \bibinfo{year}{{2014}}\natexlab{}.
\newblock \bibinfo{title}{{SNAP Datasets: Stanford Large Network Dataset
  Collection}}.
\newblock \bibinfo{howpublished}{\url{http://snap.stanford.edu/data}}.
\newblock


\bibitem[Kempe et~al\mbox{.}(2003)]%
        {kempe2003maximizing}
\bibfield{author}{\bibinfo{person}{David Kempe}, \bibinfo{person}{Jon
  Kleinberg}, {and} \bibinfo{person}{{\'E}va Tardos}.}
  \bibinfo{year}{{2003}}\natexlab{}.
\newblock \showarticletitle{{Maximizing the spread of influence through a
  social network}}. In \bibinfo{booktitle}{\emph{{ACM SIGKDD International
  Conference on Knowledge Discovery and Data Mining}}}.
  \bibinfo{publisher}{{ACM}}, \bibinfo{address}{{New York, NY, USA}},
  \bibinfo{pages}{{137--146}}.
\newblock


\bibitem[Khajehnejad et~al\mbox{.}(2020)]%
        {khajehnejad2020adversarial}
\bibfield{author}{\bibinfo{person}{Moein Khajehnejad},
  \bibinfo{person}{Ahmad~Asgharian Rezaei}, \bibinfo{person}{Mahmoudreza
  Babaei}, \bibinfo{person}{Jessica Hoffmann}, \bibinfo{person}{Mahdi Jalili},
  {and} \bibinfo{person}{Adrian Weller}.} \bibinfo{year}{{2020}}\natexlab{}.
\newblock \bibinfo{title}{{Adversarial graph embeddings for fair influence
  maximization over social networks}}.
\newblock
\newblock
\newblock
\shownote{{arXiv preprint arXiv:2005.04074}}.


\bibitem[{Konotopska, Kateryna and Iacca, Giovanni}(2021)]%
        {konotopska2021graph}
\bibfield{author}{\bibinfo{person}{{Konotopska, Kateryna and Iacca,
  Giovanni}}.} \bibinfo{year}{{2021}}\natexlab{}.
\newblock \showarticletitle{{Graph-aware evolutionary algorithms for influence
  maximization}}. In \bibinfo{booktitle}{\emph{{Genetic and Evolutionary
  Computation Conference Companion}}}. \bibinfo{publisher}{{ACM}},
  \bibinfo{address}{{New York, NY, USA}}, \bibinfo{pages}{{1467--1475}}.
\newblock


\bibitem[{Lee, Jong-Ryul and Chung, Chin-Wan}(2014)]%
        {lee2014fast}
\bibfield{author}{\bibinfo{person}{{Lee, Jong-Ryul and Chung, Chin-Wan}}.}
  \bibinfo{year}{{2014}}\natexlab{}.
\newblock \showarticletitle{{A fast approximation for influence maximization in
  large social networks}}. In \bibinfo{booktitle}{\emph{{International
  Conference on World Wide Web}}}. \bibinfo{publisher}{{ACM}},
  \bibinfo{address}{{New York, NY, USA}}, \bibinfo{pages}{{1157--1162}}.
\newblock


\bibitem[Leskovec et~al\mbox{.}(2007)]%
        {leskovec2007cost}
\bibfield{author}{\bibinfo{person}{Jure Leskovec}, \bibinfo{person}{Andreas
  Krause}, \bibinfo{person}{Carlos Guestrin}, \bibinfo{person}{Christos
  Faloutsos}, \bibinfo{person}{Jeanne VanBriesen}, {and}
  \bibinfo{person}{Natalie Glance}.} \bibinfo{year}{{2007}}\natexlab{}.
\newblock \showarticletitle{{Cost-effective outbreak detection in networks}}.
  In \bibinfo{booktitle}{\emph{{ACM SIGKDD International Conference on
  Knowledge Discovery and Data Mining}}}. \bibinfo{publisher}{{ACM}},
  \bibinfo{address}{{New York, NY, USA}}, \bibinfo{pages}{{420--429}}.
\newblock


\bibitem[{Leskovec, Jure and Huttenlocher, Daniel and Kleinberg, Jon}(2010)]%
        {leskovec2010signed}
\bibfield{author}{\bibinfo{person}{{Leskovec, Jure and Huttenlocher, Daniel and
  Kleinberg, Jon}}.} \bibinfo{year}{{2010}}\natexlab{}.
\newblock \showarticletitle{{Signed networks in social media}}. In
  \bibinfo{booktitle}{\emph{{SIGCHI Conference on Human Factors in Computing
  Systems}}}. \bibinfo{publisher}{{ACM}}, \bibinfo{address}{{New York, NY,
  USA}}, \bibinfo{pages}{{1361--1370}}.
\newblock


\bibitem[{Leskovec, Jure and Kleinberg, Jon and Faloutsos, Christos}(2007)]%
        {leskovec2007graph}
\bibfield{author}{\bibinfo{person}{{Leskovec, Jure and Kleinberg, Jon and
  Faloutsos, Christos}}.} \bibinfo{year}{{2007}}\natexlab{}.
\newblock \showarticletitle{{Graph evolution: Densification and shrinking
  diameters}}.
\newblock \bibinfo{journal}{\emph{{ACM Transactions on Knowledge Discovery from
  Data}}} \bibinfo{volume}{{1}}, \bibinfo{number}{{1}}
  (\bibinfo{year}{{2007}}), \bibinfo{pages}{{2--es}}.
\newblock


\bibitem[{Leskovec, Jure and Mcauley, Julian}(2012)]%
        {leskovec2012learning}
\bibfield{author}{\bibinfo{person}{{Leskovec, Jure and Mcauley, Julian}}.}
  \bibinfo{year}{{2012}}\natexlab{}.
\newblock \showarticletitle{{Learning to discover social circles in ego
  networks}}.
\newblock \bibinfo{journal}{\emph{{Advances in Neural Information Processing
  Systems}}}  \bibinfo{volume}{{25}} (\bibinfo{year}{{2012}}),
  \bibinfo{numpages}{{9}}~pages.
\newblock


\bibitem[Li et~al\mbox{.}(2023)]%
        {li2022piano}
\bibfield{author}{\bibinfo{person}{Hui Li}, \bibinfo{person}{Mengting Xu},
  \bibinfo{person}{Sourav~S Bhowmick}, \bibinfo{person}{Joty~Shafiq Rayhan},
  \bibinfo{person}{Changsheng Sun}, {and} \bibinfo{person}{Jiangtao Cui}.}
  \bibinfo{year}{{2023}}\natexlab{}.
\newblock \showarticletitle{{PIANO: Influence maximization meets deep
  reinforcement learning}}.
\newblock \bibinfo{journal}{\emph{{IEEE Transactions on Computational Social
  Systems}}} \bibinfo{volume}{{10}}, \bibinfo{number}{{3}}
  (\bibinfo{year}{{2023}}), \bibinfo{pages}{{1288--1300}}.
\newblock


\bibitem[Ling et~al\mbox{.}(2023)]%
        {ling2023deep}
\bibfield{author}{\bibinfo{person}{Chen Ling}, \bibinfo{person}{Junji Jiang},
  \bibinfo{person}{Junxiang Wang}, \bibinfo{person}{My~T Thai},
  \bibinfo{person}{Renhao Xue}, \bibinfo{person}{James Song},
  \bibinfo{person}{Meikang Qiu}, {and} \bibinfo{person}{Liang Zhao}.}
  \bibinfo{year}{{2023}}\natexlab{}.
\newblock \showarticletitle{{Deep graph representation learning and
  optimization for influence maximization}}. In
  \bibinfo{booktitle}{\emph{{International Conference on Machine Learning}}}.
  \bibinfo{publisher}{{PMLR}}, \bibinfo{address}{{Honolulu, HI, USA}},
  \bibinfo{pages}{{21350--21361}}.
\newblock


\bibitem[Liu et~al\mbox{.}(2012)]%
        {liu2012time}
\bibfield{author}{\bibinfo{person}{Bo Liu}, \bibinfo{person}{Gao Cong},
  \bibinfo{person}{Dong Xu}, {and} \bibinfo{person}{Yifeng Zeng}.}
  \bibinfo{year}{{2012}}\natexlab{}.
\newblock \showarticletitle{{Time constrained influence maximization in social
  networks}}. In \bibinfo{booktitle}{\emph{{IEEE International Conference on
  Data Mining}}}. \bibinfo{publisher}{{IEEE}}, \bibinfo{address}{{New York, NY,
  USA}}, \bibinfo{pages}{{439--448}}.
\newblock


\bibitem[{Lotf, Jalil Jabari and Azgomi, Mohammad Abdollahi and Dishabi,
  Mohammad Reza Ebrahimi}(2022)]%
        {lotf2022improved}
\bibfield{author}{\bibinfo{person}{{Lotf, Jalil Jabari and Azgomi, Mohammad
  Abdollahi and Dishabi, Mohammad Reza Ebrahimi}}.}
  \bibinfo{year}{{2022}}\natexlab{}.
\newblock \showarticletitle{{An improved influence maximization method for
  social networks based on genetic algorithm}}.
\newblock \bibinfo{journal}{\emph{{Physica A: Statistical Mechanics and its
  Applications}}}  \bibinfo{volume}{{586}} (\bibinfo{year}{{2022}}),
  \bibinfo{pages}{{126480}}.
\newblock


\bibitem[{Ma, Lijia and Shao, Zengyang and Li, Xiaocong and Lin, Qiuzhen and
  Li, Jianqiang and Leung, Victor CM and Nandi, Asoke K}(2022)]%
        {ma2022influence}
\bibfield{author}{\bibinfo{person}{{Ma, Lijia and Shao, Zengyang and Li,
  Xiaocong and Lin, Qiuzhen and Li, Jianqiang and Leung, Victor CM and Nandi,
  Asoke K}}.} \bibinfo{year}{{2022}}\natexlab{}.
\newblock \showarticletitle{{Influence maximization in complex networks by
  using evolutionary deep reinforcement learning}}.
\newblock \bibinfo{journal}{\emph{{IEEE Transactions on Emerging Topics in
  Computational Intelligence}}}  \bibinfo{volume}{{7}}
  (\bibinfo{year}{{2022}}), \bibinfo{pages}{{995--1009}}.
\newblock
Issue {4}.


\bibitem[McCallum et~al\mbox{.}(2000)]%
        {mccallum2000automating}
\bibfield{author}{\bibinfo{person}{Andrew~Kachites McCallum},
  \bibinfo{person}{Kamal Nigam}, \bibinfo{person}{Jason Rennie}, {and}
  \bibinfo{person}{Kristie Seymore}.} \bibinfo{year}{{2000}}\natexlab{}.
\newblock \showarticletitle{{Automating the construction of internet portals
  with machine learning}}.
\newblock \bibinfo{journal}{\emph{{Information Retrieval}}}
  \bibinfo{volume}{{3}} (\bibinfo{year}{{2000}}), \bibinfo{pages}{{127--163}}.
\newblock


\bibitem[Olivares et~al\mbox{.}(2021)]%
        {olivares2021multi}
\bibfield{author}{\bibinfo{person}{Rodrigo Olivares},
  \bibinfo{person}{Francisco Mu{\~n}oz}, {and} \bibinfo{person}{Fabi{\'a}n
  Riquelme}.} \bibinfo{year}{{2021}}\natexlab{}.
\newblock \showarticletitle{{A multi-objective linear threshold influence
  spread model solved by swarm intelligence-based methods}}.
\newblock \bibinfo{journal}{\emph{{Knowledge-Based Systems}}}
  \bibinfo{volume}{{212}} (\bibinfo{year}{{2021}}), \bibinfo{pages}{{106623}}.
\newblock


\bibitem[Panagopoulos et~al\mbox{.}(2020)]%
        {panagopoulos2020multi}
\bibfield{author}{\bibinfo{person}{George Panagopoulos},
  \bibinfo{person}{Fragkiskos~D Malliaros}, {and} \bibinfo{person}{Michalis
  Vazirgiannis}.} \bibinfo{year}{{2020}}\natexlab{}.
\newblock \showarticletitle{{Multi-task learning for influence estimation and
  maximization}}.
\newblock \bibinfo{journal}{\emph{{IEEE Transactions on Knowledge and Data
  Engineering}}} \bibinfo{volume}{{34}}, \bibinfo{number}{{9}}
  (\bibinfo{year}{{2020}}), \bibinfo{pages}{{4398--4409}}.
\newblock


\bibitem[{Peixoto, Tiago P}(2020)]%
        {peixoto2020netzschleuder}
\bibfield{author}{\bibinfo{person}{{Peixoto, Tiago P}}.}
  \bibinfo{year}{2020}\natexlab{}.
\newblock \bibinfo{title}{{The Netzschleuder network catalogue and
  repository}}.
\newblock \bibinfo{howpublished}{\url{https://networks.skewed.de}}.
\newblock


\bibitem[{Perrault, Pierre and Healey, Jennifer and Wen, Zheng and Valko,
  Michal}(2020)]%
        {perrault2020budgeted}
\bibfield{author}{\bibinfo{person}{{Perrault, Pierre and Healey, Jennifer and
  Wen, Zheng and Valko, Michal}}.} \bibinfo{year}{{2020}}\natexlab{}.
\newblock \showarticletitle{{Budgeted online influence maximization}}. In
  \bibinfo{booktitle}{\emph{{International Conference on Machine Learning}}}.
  \bibinfo{publisher}{{PMLR}}, \bibinfo{address}{{virtual event}},
  \bibinfo{pages}{{7620--7631}}.
\newblock


\bibitem[Pham et~al\mbox{.}(2019)]%
        {pham2019competitive}
\bibfield{author}{\bibinfo{person}{Canh~V Pham}, \bibinfo{person}{Hieu~V
  Duong}, \bibinfo{person}{Huan~X Hoang}, {and} \bibinfo{person}{My~T Thai}.}
  \bibinfo{year}{{2019}}\natexlab{}.
\newblock \showarticletitle{{Competitive influence maximization within time and
  budget constraints in online social networks: An algorithmic approach}}.
\newblock \bibinfo{journal}{\emph{{Applied Sciences}}} \bibinfo{volume}{{9}},
  \bibinfo{number}{{11}} (\bibinfo{year}{{2019}}), \bibinfo{pages}{{2274}}.
\newblock


\bibitem[Razaghi et~al\mbox{.}(2022)]%
        {razaghi2022group}
\bibfield{author}{\bibinfo{person}{Behnam Razaghi}, \bibinfo{person}{Mehdy
  Roayaei}, {and} \bibinfo{person}{Nasrollah~Moghadam Charkari}.}
  \bibinfo{year}{{2022}}\natexlab{}.
\newblock \showarticletitle{{On the Group-Fairness-Aware Influence Maximization
  in Social Networks}}.
\newblock \bibinfo{journal}{\emph{{IEEE Transactions on Computational Social
  Systems}}} \bibinfo{volume}{{10}}, \bibinfo{number}{{6}}
  (\bibinfo{year}{{2022}}), \bibinfo{pages}{{3406--3414}}.
\newblock


\bibitem[Riquelme et~al\mbox{.}(2023)]%
        {riquelme2023depth}
\bibfield{author}{\bibinfo{person}{Fabi{\'a}n Riquelme},
  \bibinfo{person}{Francisco Mu{\~n}oz}, {and} \bibinfo{person}{Rodrigo
  Olivares}.} \bibinfo{year}{{2023}}\natexlab{}.
\newblock \showarticletitle{{A depth-based heuristic to solve the
  multi-objective influence spread problem using particle swarm optimization}}.
\newblock \bibinfo{journal}{\emph{{OPSEARCH}}}  \bibinfo{volume}{{60}}
  (\bibinfo{year}{{2023}}), \bibinfo{pages}{{1267--1285}}.
\newblock


\bibitem[Rossi and Ahmed(2015)]%
        {rossi2015network}
\bibfield{author}{\bibinfo{person}{Ryan Rossi} {and} \bibinfo{person}{Nesreen
  Ahmed}.} \bibinfo{year}{{2015}}\natexlab{}.
\newblock \showarticletitle{{The network data repository with interactive graph
  analytics and visualization}}. In \bibinfo{booktitle}{\emph{{AAAI Conference
  on Artificial Intelligence}}}, Vol.~\bibinfo{volume}{{29}}.
  \bibinfo{publisher}{{Association for the Advancement of Artificial
  Intelligence}}, \bibinfo{address}{{Washington DC, USA}},
  \bibinfo{pages}{{4292--4293}}.
\newblock


\bibitem[Rui et~al\mbox{.}(2023)]%
        {rui2023scalable}
\bibfield{author}{\bibinfo{person}{Xiaobin Rui}, \bibinfo{person}{Zhixiao
  Wang}, \bibinfo{person}{Jiayu Zhao}, \bibinfo{person}{Lichao Sun}, {and}
  \bibinfo{person}{Wei Chen}.} \bibinfo{year}{{2023}}\natexlab{}.
\newblock \showarticletitle{{Scalable Fair Influence Maximization}}.
\newblock \bibinfo{journal}{\emph{{Advances in Neural Information Processing
  Systems}}}  \bibinfo{volume}{{1}} (\bibinfo{year}{{2023}}),
  \bibinfo{numpages}{{12}}~pages.
\newblock


\bibitem[{Shang, Ke and Ishibuchi, Hisao and He, Linjun and Pang, Lie
  Meng}(2020)]%
        {shang2020survey}
\bibfield{author}{\bibinfo{person}{{Shang, Ke and Ishibuchi, Hisao and He,
  Linjun and Pang, Lie Meng}}.} \bibinfo{year}{{2020}}\natexlab{}.
\newblock \showarticletitle{{A survey on the hypervolume indicator in
  evolutionary multiobjective optimization}}.
\newblock \bibinfo{journal}{\emph{{IEEE Transactions on Evolutionary
  Computation}}} \bibinfo{volume}{{25}}, \bibinfo{number}{{1}}
  (\bibinfo{year}{{2020}}), \bibinfo{pages}{{1--20}}.
\newblock


\bibitem[Stoica et~al\mbox{.}(2020)]%
        {stoica2020seeding}
\bibfield{author}{\bibinfo{person}{Ana-Andreea Stoica},
  \bibinfo{person}{Jessy~Xinyi Han}, {and} \bibinfo{person}{Augustin
  Chaintreau}.} \bibinfo{year}{{2020}}\natexlab{}.
\newblock \showarticletitle{{Seeding network influence in biased networks and
  the benefits of diversity}}. In \bibinfo{booktitle}{\emph{{The Web
  Conference}}}. \bibinfo{publisher}{{ACM}}, \bibinfo{address}{{New York, NY,
  USA}}, \bibinfo{pages}{{2089--2098}}.
\newblock


\bibitem[Tang et~al\mbox{.}(2018)]%
        {tang2018online}
\bibfield{author}{\bibinfo{person}{Jing Tang}, \bibinfo{person}{Xueyan Tang},
  \bibinfo{person}{Xiaokui Xiao}, {and} \bibinfo{person}{Junsong Yuan}.}
  \bibinfo{year}{{2018}}\natexlab{}.
\newblock \showarticletitle{{Online processing algorithms for influence
  maximization}}. In \bibinfo{booktitle}{\emph{{ACM SIGMOD International
  Conference on Management of Data}}}. \bibinfo{publisher}{{ACM}},
  \bibinfo{address}{{New York, NY, USA}}, \bibinfo{pages}{{991--1005}}.
\newblock


\bibitem[Tang et~al\mbox{.}(2015)]%
        {tang2015influence}
\bibfield{author}{\bibinfo{person}{Youze Tang}, \bibinfo{person}{Yanchen Shi},
  {and} \bibinfo{person}{Xiaokui Xiao}.} \bibinfo{year}{{2015}}\natexlab{}.
\newblock \showarticletitle{{Influence maximization in near-linear time: A
  martingale approach}}. In \bibinfo{booktitle}{\emph{{ACM SIGMOD International
  Conference on Management of Data}}}. \bibinfo{publisher}{{ACM}},
  \bibinfo{address}{{New York, NY, USA}}, \bibinfo{pages}{{1539--1554}}.
\newblock


\bibitem[Tong et~al\mbox{.}(2020)]%
        {tong2020time}
\bibfield{author}{\bibinfo{person}{Guangmo Tong}, \bibinfo{person}{Ruiqi Wang},
  \bibinfo{person}{Zheng Dong}, {and} \bibinfo{person}{Xiang Li}.}
  \bibinfo{year}{{2020}}\natexlab{}.
\newblock \showarticletitle{{Time-constrained adaptive influence
  maximization}}.
\newblock \bibinfo{journal}{\emph{{IEEE Transactions on Computational Social
  Systems}}} \bibinfo{volume}{{8}}, \bibinfo{number}{{1}}
  (\bibinfo{year}{{2020}}), \bibinfo{pages}{{33--44}}.
\newblock


\bibitem[{Traag, Vincent A and Waltman, Ludo and Van Eck, Nees Jan}(2019)]%
        {traag2019louvain}
\bibfield{author}{\bibinfo{person}{{Traag, Vincent A and Waltman, Ludo and Van
  Eck, Nees Jan}}.} \bibinfo{year}{{2019}}\natexlab{}.
\newblock \showarticletitle{{From Louvain to Leiden: guaranteeing
  well-connected communities}}.
\newblock \bibinfo{journal}{\emph{{Scientific reports}}} \bibinfo{volume}{{9}},
  \bibinfo{number}{{1}} (\bibinfo{year}{{2019}}), \bibinfo{pages}{{5233}}.
\newblock


\bibitem[Tsang et~al\mbox{.}(2019)]%
        {tsang2019group}
\bibfield{author}{\bibinfo{person}{Alan Tsang}, \bibinfo{person}{Bryan Wilder},
  \bibinfo{person}{Eric Rice}, \bibinfo{person}{Milind Tambe}, {and}
  \bibinfo{person}{Yair Zick}.} \bibinfo{year}{{2019}}\natexlab{}.
\newblock \bibinfo{title}{{Group-fairness in influence maximization}}.
\newblock
\newblock
\newblock
\shownote{{arXiv preprint arXiv:1903.00967}}.


\bibitem[{Tukey, John W}(1949)]%
        {tukey1949comparing}
\bibfield{author}{\bibinfo{person}{{Tukey, John W}}.}
  \bibinfo{year}{{1949}}\natexlab{}.
\newblock \showarticletitle{{Comparing individual means in the analysis of
  variance}}.
\newblock \bibinfo{journal}{\emph{{Biometrics}}} \bibinfo{volume}{{5}},
  \bibinfo{number}{{2}} (\bibinfo{year}{{1949}}), \bibinfo{pages}{{99--114}}.
\newblock


\bibitem[Wang et~al\mbox{.}(2016)]%
        {wang2016maximizing}
\bibfield{author}{\bibinfo{person}{Xiaojie Wang}, \bibinfo{person}{Xue Zhang},
  \bibinfo{person}{Chengli Zhao}, {and} \bibinfo{person}{Dongyun Yi}.}
  \bibinfo{year}{{2016}}\natexlab{}.
\newblock \showarticletitle{{Maximizing the spread of influence via generalized
  degree discount}}.
\newblock \bibinfo{journal}{\emph{{PloS ONE}}} \bibinfo{volume}{{11}},
  \bibinfo{number}{{10}} (\bibinfo{year}{{2016}}), \bibinfo{pages}{{e0164393}}.
\newblock


\bibitem[{Watts, Duncan J and Strogatz, Steven H}(1998)]%
        {watts1998collective}
\bibfield{author}{\bibinfo{person}{{Watts, Duncan J and Strogatz, Steven H}}.}
  \bibinfo{year}{{1998}}\natexlab{}.
\newblock \showarticletitle{{Collective dynamics of
  ‘small-world’networks}}.
\newblock \bibinfo{journal}{\emph{{Nature}}} \bibinfo{volume}{{393}},
  \bibinfo{number}{{6684}} (\bibinfo{year}{{1998}}),
  \bibinfo{pages}{{440--442}}.
\newblock


\end{thebibliography}


\end{document}